%% file: main.tex
\documentclass[twoside,11pt]{article}

\usepackage{blindtext}

\usepackage[abbrvbib, preprint]{jmlr2e}

\usepackage{xcolor}
\usepackage{mathtools, amssymb}
\usepackage{stmaryrd}
\usepackage{bbm}
\usepackage{hyperref}
\usepackage{booktabs}
\usepackage{longtable}
\usepackage{array}
\usepackage{algorithm}
\usepackage{algorithmic}
\usepackage[inline]{enumitem}
\usepackage{float}

\newcommand{\category}[1]{\vspace{0.5em} \noindent \textbf{#1} \vspace{0.5em}}

\usepackage{lastpage}
\jmlrheading{23}{2023}{1-\pageref{LastPage}}{1/21; Revised 5/22}{9/22}{21-0000}{Rubén Ballester and Carles Casacuberta and Sergio Escalera}

\ShortHeadings{TDA for Neural Network Analysis}{TDA for Neural Network Analysis}
\firstpageno{1}

\begin{document}

\title{Topological Data Analysis for Neural Network Analysis: A~Comprehensive Survey}

\author{\name Rubén Ballester \email ruben.ballester@ub.edu \\
        \name Carles Casacuberta \email carles.casacuberta@ub.edu \\
       \addr Departament de Matemàtiques i Informàtica\\
             Universitat de Barcelona\\
             Gran Via de les Corts Catalanes, 585, 08007 Barcelona, Catalonia, Spain \\
       \AND
       \name Sergio Escalera \email sescalera@ub.edu \\
       \addr Departament de Matemàtiques i Informàtica\\
             Universitat de Barcelona\\
             Gran Via de les Corts Catalanes, 585, 08007 Barcelona, Catalonia, Spain \\
             Computer Vision Center\\
             Edifici O, Campus UAB, 08193 Bellaterra, Catalonia, Spain}

\editor{My editor} 

\maketitle

\begin{abstract}
This survey provides a comprehensive exploration of applications of Topological Data Analysis (TDA) within neural network analysis. Using TDA tools such as persistent homology and Mapper, we delve into the intricate structures and behaviors of neural networks and their datasets. We discuss different strategies to obtain topological information from data and neural networks by means of TDA. Additionally, we review how  topological information can be leveraged to analyze properties of neural networks, such as their generalization capacity or expressivity. We explore practical implications of deep learning, specifically focusing on areas like adversarial detection and model selection. Our survey organizes the examined works into four broad domains:
\begin{enumerate*}
    \item Characterization of neural network architectures;
    \item Analysis of decision regions and boundaries;
    \item Study of internal representations, activations, and parameters; 
    \item Exploration of training dynamics and loss functions.
\end{enumerate*}
Within each category, we discuss several articles, offering background information to aid in understanding the various methodologies. We conclude with a synthesis of key insights gained from our study, accompanied by a discussion of challenges and potential advancements in the field.
\end{abstract}


\section{Introduction}\label{scn:introduction}
Over the past few years, deep learning has consolidated its position as the most successful branch of artificial intelligence. With the continuous growth in computational capacity, neural networks have expanded in size and complexity, enabling them to effectively tackle progressively difficult problems. However, their increased capacity has made it more challenging to comprehend essential properties of the networks such as their interpretability, generalization ability, or suitability for specific problems. From both theoretical and practical standpoints, this is undesirable, especially in critical contexts where AI decisions could lead to catastrophic consequences, such as medical diagnosis~\citep{intelligent_healthcare_survey} or autonomous driving~\citep{review_ai_safety_autonomous_driving}, among others. 

Topological Data Analysis (TDA) has emerged as a subfield of algebraic topology that offers a framework for gaining insights into the \emph{shape} of data in a broad sense. Topological data analysis has found application across a wide array of experimental science disciplines, spanning from biomedicine~\citep{tda_in_biomedicine_review} to finance~\citep{tda_financial_time_series}, among numerous others. One of its most prolific areas of application is machine learning, particularly in the domain of deep learning. A basic introduction to topological machine learning can be found in~\cite{tml_survey}. 
Topological data analysis, specifically homology, persistent homology and Mapper, has been used to analyze various aspects of neural networks. Broadly, these aspects can be categorized in the following four groups:
\begin{enumerate*}
    \item Structure of neural networks;
    \item Input and output spaces;
    \item Internal representations and activations;
    \item Training dynamics and loss functions.
\end{enumerate*}

\begin{figure}[ht]
\centering
\includegraphics[width=0.98\textwidth]{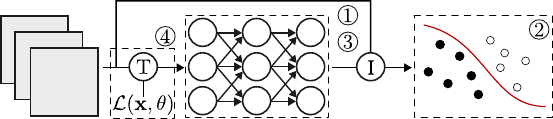}
\caption{Diagram showing the usual lifecycle of a neural network $\mathcal{N}$. First, an architecture $a(\mathcal{N})$ is selected based on the task to be solved. This architecture is independent of the learned parameters $\theta(\mathcal{N})$ or the specific input data used to train or test the network, denoted $\mathcal{D}_\text{train}$ and $\mathcal{D}_\text{test}$, respectively. Second, the architecture is trained ($\text{T}$) using a specific training algorithm $\mathcal{A}$, which generally minimizes the empirical risk of a loss function $\mathcal{L}$ evaluated on the training dataset $\mathcal{D}_\text{train}$. Once the network is trained, inference ($\text{I}$) is performed with data coming from the same distribution $\mathbb P$ from which the training data were sampled. For trained neural networks, input and output spaces gather several interesting structures, such as decision regions and boundaries for classification problems or latent spaces for generative models, among others. Each dashed box is related to one of the categories, labeled $1$ to~$4$, in which topological data analysis has been used to analyze neural networks. The categories are the following: (1)~Structure of the neural network; (2)~Input and output spaces; (3)~Internal representations and activations; (4)~Training dynamics and loss functions. The leftmost box contains the training part of the lifecycle of a neural network and is related to category~4. The central box contains the neural network and is concerned with categories 1 and~3. The rightmost box contains the decision regions and boundaries of the output space of a neural network after training, which are related to category~2.}
\label{fig:categories_of_dl_analysis_using_tda}
\end{figure}

Figure~\ref{fig:categories_of_dl_analysis_using_tda} visually delineates these four categories. The first category involves examining unweighted graphs associated with neural networks and their properties, such as depth, layer widths, and graph topology, among others. The second category encompasses the analysis of neural network input and output spaces, including decision regions and boundaries for classification problems, as well as the study of latent spaces for generative models. The third category, which currently holds the largest number of contributions in the literature, focuses on the analysis of hidden and output neurons in a broad sense. Lastly, the fourth category involves the analysis of neural network training procedures, including the study of loss functions.

Many components studied within the preceding categories play a pivotal role in understanding some of the fundamental traits of deep learning, such as interpretability or the generalization capacity of neural networks. Moreover, these elements inherently exhibit geometrical and topological characteristics, rendering them exceptionally suitable for the application of topological data analysis methods.

\subsection{Contribution}

In this work, we offer a comprehensive overview of applications of topological data analysis in analyzing neural networks across the aforementioned four categories. However, we have omitted many relevant works related to the general use of topological data analysis in deep learning. For example, we omit work on the branch of Topological Deep Learning, which involves developing neural networks tailored for specific topological data. A recent survey on Topological Deep Learning is available in~\cite{survey_topological_neural_networks}. We have also omitted the majority of applications using topological data analysis to construct loss functions, since our focus remains on the analysis of neural networks rather than their enhancement through the imposition of specific topological structures on data, unrelated to the particular machine learning algorithm used for the task. Given the substantial volume of papers analyzed, our discussion is limited to peer-reviewed papers, with occasional exceptions made for relevant and credible sources.

This survey aims to be self-contained and approachable for readers unfamiliar with topological data analysis or deep learning.
However, it is advisable that readers have a background in at least one of these areas. For mathematicians interested in how topology can be applied to study modern AI and deep learning systems, we recommend reading the first chapter of the book by~\cite{mathematical_aspects_of_deep_learning}, that provides a concise introduction to deep learning from a mathematical viewpoint. For machine learning scientists curious about how advanced topological methods can enhance the understanding of deep learning systems, we recommend the survey on Topological Machine Learning by~\cite{tml_survey}, where topological data analysis is introduced in a friendly manner within the scope of machine learning.

\subsection{Outline}

In Section~\ref{scn:preeliminaries} we introduce notation, definitions, and basic results of topological data analysis and deep learning needed to follow the survey. Section~\ref{scn:analysis_dnn_using_tda} constitutes the core content of our survey, structured into the four aforementioned categories. Finally, Section~\ref{scn:challenges_future_directions_conclusions} addresses limitations, challenges, and potential future directions concerning the application of TDA in deep learning up to the time of publication of this survey.

\tableofcontents

\vspace*{0.3cm}

\section{Preliminaries}\label{scn:preeliminaries}

Preliminary definitions and results for the deep learning part are primarily drawn from the first chapter of the book by~\cite{mathematical_aspects_of_deep_learning}. Regarding the topological data analysis part, most of the content is sourced from the book by~\cite{edelsbrunner2022computational} and the survey conducted by~\cite{tml_survey}.

\subsection{Notation}

We denote by $\mathbb{N}$, $\mathbb{Z}$, $\mathbb{R}$, and $\bar{\mathbb R}$ respectively the set of natural numbers (including zero), the ring of integers, the field of real numbers, and the extended real line with a point at infinity. For $n\ge 2$, we 
abbreviate $\{1,\hdots, n\}$ as~$[n]$. Given a set $S$, its power set, i.e., the set of its subsets, is denoted by~$\mathcal{P}(S)$. The indicator function on a set $A$ is written as $\mathbbm{1}_A$ and it is defined as $\mathbbm 1_A(x)=1$ if $x\in A$ and $\mathbbm 1_A(x)=0$ if $x\not\in A$. The Kronecker delta $\delta_{i,j}$ is defined as $\delta_{i,j}=1$ if $i=j$ and $\delta_{i,j}=0$ if $i\neq j$.

The set of matrices of size $m\times n$ with coefficients in 
a ring $S$ (usually a field) is denoted by $M_{m,n}(S)$. The $(i,j)$ entry of a matrix $M$ is written as $M_{i,j}$. For a square matrix~$M$, its trace is denoted by $\text{tr}(M)$. Given a vector $v\in\mathbb R^d$, we denote its $i$-th component by $v_i$.

We denote by $\mathcal{M}(\mathcal{X}, \mathcal{Y})$ the set of all measurable functions from $\mathcal{X}$ to $\mathcal{Y}$, where $\mathcal{X}$ and $\mathcal{Y}$ are measure spaces.  Given a function $f$, we denote its domain and image as $\text{Dom}(f)$ and $\text{Im}(f)$, respectively, and we denote its graph by $G(f)= \{(x, f(x)):x\in\text{Dom}(f)\}$. 

The \emph{support} of a function $f\colon X\to\mathbb{R}$, denoted $\text{supp}(f)$, is the set of elements of $X$ that are not sent to zero, or most commonly the closure of this set if $X$ is a topological space. Similarly, for a probability distribution $\mathbb P$ associated with a random variable~$X$, the support 
$\text{supp}(\mathbb P)$
is the smallest closed set $S$ of real numbers such that $\mathbb P(X\in S)=1$. We also call support of $\mathbb P$ the set of elements of the sample space mapped to $\text{supp}(\mathbb P)$ by~$X$. The letter $\mathbb E$ is used to denote expectation of a random variable.

For a graph $G$, we denote by $V(G)$ and $E(G)$ the sets of vertices and edges of~$G$, respectively. By a \emph{weighted graph} we refer to an undirected graph $G$ whose vertices and edges are weighted by functions $w_V\colon V(G)\to\mathbb R$ and $w_E\colon E(G)\to\mathbb R$, and we use the notation
$\left(G, w_V, w_E\right)$. A bipartite graph is denoted by $G=(V_1, V_2, E)$, where $V_1$ and $V_2$ are disjoint sets of vertices. 

\subsection{Deep learning}\label{scn:deep_learning_fundamentals}

Deep learning is a subfield of machine learning that deals with (deep) neural networks to solve a variety of computational tasks. Some computational tasks include, but are not limited to, classification, regression, and synthetic data generation. These three tasks are central in this survey.

Classification and regression are particular examples of prediction tasks. In prediction tasks, we have data lying on the Cartesian product of two measure spaces $\mathcal{X}$ and $\mathcal{Y}$, where elements of $\mathcal{Y}$ are seen as \emph{labels} of elements of~$\mathcal{X}$. Such data are distributed according to a distribution $\mathbb{P}_{(X, Y)}$ with marginals $\mathbb P_X$ and $\mathbb P_Y$ for the restrictions to $\mathcal{X}$ and $\mathcal Y$, respectively. In prediction tasks, we look for a function $f\in\mathcal{M}\left(\mathcal{X}, \mathcal{Y}\right)$ that, given a value $x\in\mathcal{X}$, outputs a ``good'' label prediction $y\in\mathcal{Y}$ according to some quality criterion. 
To precisely quantify the concept of quality, a loss function 
\[
\mathcal{L}\colon \mathcal{M}(\mathcal{X}, \mathcal{Y})\times\mathcal{X}\times\mathcal{Y}\longrightarrow\mathbb R
\] 
is typically employed. A~loss function indicates the degree of inaccuracy in predictions relative to a data point $(x,y)\in\mathcal{X}\times\mathcal{Y}$ sampled from the data distribution ---where a lower value signifies greater quality.
Thus, the objective is to find a function $f$ that minimizes the expected loss $\mathbb E\left[\mathcal{L}\left(f, X, Y\right)\right]$, called \emph{risk} of $f$ and denoted by $\mathcal{R}(f)$. Due to the hardness of finding such $f$ (if it exists) in the huge space $\mathcal{M}\left(\mathcal{X}, \mathcal{Y}\right)$, the problem is often reduced to find a function $f$ whose risk is arbitrarily near to the infimum of risks in a smaller set of functions called \emph{hypothesis set} and denoted by~$\mathcal{F}$. 

One of the main difficulties in this approach is that the data distribution $\mathbb{P}_{(X,Y)}$ is often unknown, and we only have access to a finite sample $\mathcal{D}=\{\left(x_i, y_i\right)\}_{i=1}^m$
from it, that we assume 
independent and identically distributed (i.i.d.) with distribution
$\mathbb P_{(X,Y)}^m$.
Therefore, instead of trying to minimize directly the risk $\mathcal{R}(f)$, one tries to minimize the empirical risk $\widehat{\mathcal{R}}_{\mathcal{D}_\text{train}}(f)$ for a subset $\mathcal{D}_\text{train}\subseteq\mathcal{D}$, called \emph{training dataset}, given by
\begin{equation}
\widehat{\mathcal{R}}_{\mathcal{D}_\text{train}}(f)= \frac{1}{m}\sum_{i=1}^m \mathcal{L}\left(f,x_i, y_i\right),
\end{equation}
with the hope that minimizing the empirical risk also reduces the real risk. In some situations, it is convenient to minimize the empirical risk of another related loss $\mathcal{L}^\text{surr}$ that is not the original loss function $\mathcal{L}$ and that can depend on a surrogate function $f^{\text{surr}}$ depending on~$f$, instead of using $f$ directly. This happens, for example, when we use minimization algorithms that rely on the differentiability of the loss function, yet our function $\mathcal{L}$ is not differentiable or its gradient is zero. 

In deep learning, one formally sees the process of obtaining a neural network $\mathcal{N}$ given data distributed according to $\mathbb P_{(X,Y)}$ as a map
\begin{equation}
\mathcal{A}\colon\bigcup_{m\in\mathbb N} \left(\mathcal{X}\times \mathcal{Y}\right)^m\longrightarrow\mathcal F
\end{equation}
that takes as input a training (ordered) set $\mathcal{D}=\{\left(x_i, y_i\right)\}_{i=1}^m$ and provides a neural network $\mathcal{N}\in\mathcal{F}$ that minimizes the empirical risk for the original or surrogate loss on $\mathcal{D}_\text{train}$. This map is called \emph{training algorithm}. Training algorithms are generally iterative minimization methods based on gradient descent. 

Regression and classification are the two modalities studied for prediction tasks in this survey. In both problems, we assume that there is an unknown function $f\colon\mathcal X\to\mathcal Y$ assigning, to each $x\in\mathcal{X}$, a \textit{true value} $y=f(x)$, and that the data distribution $\mathbb P_{(X,Y)}$ 
has the property that 
$G(f)\subseteq\text{supp}\left(\mathbb P_{(X,Y)}\right)$. In this scenario, the goal of a neural network is, for each value $x\in \mathcal{X}$, to output the corresponding value~$f(x)$. 

We say that a prediction task is a \emph{regression task} if $\mathcal{X}$ is a Euclidean space $\mathbb R^d$ for some $d\in\mathbb N_{> 0}$ and $\mathcal{Y}=\mathbb R$, and we say that a prediction task is a \emph{classification task} if $\mathcal{X}$ is a Euclidean space $\mathbb R^d$ and $\mathcal{Y}$ is a finite set. In the latter case we assume, without loss of generality, that $\mathcal{Y}=[k]$ with $k\in\mathbb N_{\geq 2}$ and that $\text{supp}\left(\mathbb P_Y\right)=\mathcal{Y}$.
If $k=2$, then we call it a \emph{binary} classification task. For a broader introduction to machine learning tasks, we refer the reader to~\citet[Section~5.1.1]{deep_learning_goodfellow}.

Generative tasks are not as straightforward to define as prediction tasks, and formal definitions can vary depending on the problem or the solving method. 
In generative tasks, the purpose is to generate synthetic data that are indistinguishable from source data. We assume that source data live in an ambient space $\mathcal{X}$ and follow a distribution $\mathbb P_X$. Some models explicitly approximate the given probability distribution, allowing to directly 
sample from it, while other models learn a mechanism to generate new examples without explicitly describing the data distribution. 
We refer to~\citet[Section~1.2.1]{prince2023understanding} for a detailed explanation of generative models.

In this survey we mainly consider \emph{fully connected feedforward neural networks} (FCFNN). FCFNNs follow a basic structure in which most of the current, more complex, models, can be expressed. A FCFNN is a parameterized function $\mathcal{N}$ consisting of:
\begin{itemize}
\item[(i)]
An \emph{architecture} $a(\mathcal{N})=(N, \varphi)$, where $N=(N_0, \hdots, N_L)\in \mathbb N^{L+1}$ and $\varphi=\left(\varphi^{(l)}\right)_{l=1}^L$ is a tuple of differentiable functions $\varphi^{(l)}\colon\mathbb R\to\mathbb R$. 
Here differentiability is meant except perhaps on a set of zero measure. The number $L$ counts the layers of the network, while $N_0$, $N_L$, and $N_l$ for $l\in[L-1]$ are the numbers of neurons of the input, output, and $l$-th hidden layers, respectively, and $\varphi^{(l)}$ is the \emph{activation function} of layer~$l$.
\item[(ii)] A set of parameters $\theta(\mathcal{N})=\left(\theta^{(l)}\right)_{l=1}^L$, where
$\theta^{(l)}=\left(W^{(l)}, b^{(l)}\right)$ 
are the parameters of layer~$l$, belonging to $\mathbb R^{ N_l \times N_{l-1}}\times \mathbb R^{N_l}$ for $l\in [L]$ and known as \emph{weights} and \emph{biases}, respectively.
\end{itemize}

\begin{figure}[ht]
\centering
\includegraphics[width=0.45\textwidth]{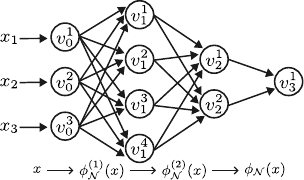}
\caption{A graphical representation of a fully connected feedforward neural network $\mathcal{N}$ with $L=3$ and $N=(3,4,2,1)$.  Each FCFNN can be represented as a sequence of sets of vertices, called layers, with vertices of the layer $l$ connected with the vertices of the layer $l-1$, for $l\in[L]$. Given an input $x\in\mathbb R^{N_0}=\mathbb R^{3}$, the input values $x_i$ are associated with the vertices $v_0^i$ of the first (input) layer and then transformed sequentially by a set of maps. In this representation, each edge indicates that the value of the source vertex is used for the computation of the value of the target vertex. Values for vertices are computed sequentially from the first layer to the last. Each transformation from layer $l-1$ to layer~$l$, made by the corresponding function~$\phi_\mathcal{N}^{(l)}$, is a composite of an affine transformation $\bar{\phi}_\mathcal{N}^{(l)}$ and the activation function $\varphi^{(l)}$ applied elementwise to all the outputs of $\bar{\phi}_\mathcal{N}^{(l)}$.}
\label{fig:neural_network_explained}
\end{figure}

A FCFNN induces a directed graph and a function depending on it. The directed graph defined by the architecture, denoted by $G\left(\mathcal{N}\right)$, is given by $L+1$ disjoint ordered sets $V_l=\{v_l^1,\hdots, v_l^{N_l}\}$, $l=0,\dots,L$, with the property that every vertex in $V_l$ has an edge pointing to it from all the vertices in $V_{l-1}$, for $l\in[L]$. 
The vertices of this graph are called \emph{neurons}. 
The function $\phi_\mathcal{N}\colon \mathbb R^{N_0}\to\mathbb R^{N_L}$ defined on top of the directed graph is given by the recursive formula
\begin{equation}
\begin{split}
    \phi_\mathcal{N}(x) &= \phi_\mathcal{N}^{(L)}(x),\\[0.2cm]
    \phi_\mathcal{N}^{(l)}(x) &=  \varphi^{(l)}\left(\bar{\phi}_\mathcal{N}^{(l)}(x)\right)\text{ for }l\in [L],\\[0.2cm]
    \bar{\phi}
    _\mathcal{N}^{(l)}(x) &= \begin{cases}
         W^{(1)}x+b^{(1)}&\text{if }l=1,\\[0.1cm]
         W^{(l)}\phi_\mathcal{N}^{(l-1)}(x)+b^{(l)}&\text{if }l\in\{2,\hdots, L\},
    \end{cases}
    \end{split}
\end{equation}
where each component $x_i$ of $x$ is associated with the vertex~$v_0^i$, and each intermediate function value $\phi^{(l)}_\mathcal{N}(x)_i$ is associated with the vertex~$v_l^i$. Given an input value~$x$, we say that the values $\phi^{(l)}_\mathcal{N}(x)_i$ are \emph{activations} of the vertices~$v_l^i$. A visual example of a FCFNN is shown in Figure~\ref{fig:neural_network_explained}.

Given a specific tuple $a=\left(N, \varphi\right)$ representing an architecture for a FCFNN and a subset $\Theta\subseteq \prod_{l\in[L]}\mathbb R^{\mathbb N_l \times \mathbb N_{l-1}}\times \mathbb R^{N_l}$, we denote the hypothesis set of all the neural networks represented by the architecture $a$ with parameters in $\Theta$ as
\[
\mathcal{F}_{a,\Theta}=\left\{\mathcal{N}:a(\mathcal{N})=a,\, \theta(\mathcal{N})\in\Theta\right\}.
\]
If $\Theta$ is the space of all possible parameters, we simply write $\mathcal{F}_a$. Also, when it is clear that we speak about neural networks in a specific hypothesis set $\mathcal{F}_{a,\Theta}$, we denote the neural network with architecture $a$ and parameters $\theta\in\Theta$ as $\mathcal{N}_\theta$.

Note that, fixing a dataset $\mathcal{D}$, an architecture $a$, and a set of possible parameters $\Theta$, the composition of the empirical risk with the neural network functions in $\mathcal{F}_{a, \Theta}$ can be seen as a function 
$\widehat{\mathcal{R}}_{\mathcal{D}_\text{train}}(\theta)$ of the parameters~$\theta$. Usually, training algorithms for neural networks, given an initial set of parameters $\theta^{(0)}\in\Theta$, generate a discrete weight trajectory $\left(\theta^{(i)}\right)_{i=1}^N$ by minimizing the empirical risk $\widehat{\mathcal{R}}_{\mathcal{D}_\text{train}}(\theta)$ iteratively with respect to the parameters $\theta$ until some stopping criteria are satisfied. The output of the algorithm is one of the weights $\theta^{(i)}$ generated in the trajectory, usually the last one, $\theta^{(N)}$. In this survey, we assume that neural networks are trained in this way.

A special type of FCFNN is a \emph{convolutional neural network} (CNN), which is a FCFNN with a specific architecture designed to work with data that have a grid-like structure, especially images. A~CNN is one of the most common type of neural networks used in computer vision. CNNs implement two functions between layers that impose restrictions on the set of weights and on the activation functions: convolutional layers and pooling operators. A \emph{convolutional layer} is a layer whose outputs are the result of performing convolutions between the values associated to the neurons of its input layer rearranged as a grid and a filter tensor that imposes a set of equalities on the coefficients of the weight matrices. In one of the most typical scenarios, a convolutional layer between layer $i-1$ and $i$ is a transformation between the activations of $V_{i-1}$ and $V_i$ in such a way $V_{i-1}$ and $V_i$ are ordered in grid structures of size $h^{(i-1)\times }w^{(i-1)}\times c^{(i-1)}$ and $h^{(i)\times }w^{(i)}\times c^{(i)}$, respectively. We say that a layer $i$ arranged in this grid structure has \emph{height} $h^{(i)}$, \emph{width} $w^{(i)}$, and $c^{i}$ \emph{channels}, where each channel is defined as the subgrid obtained by fixing one value in $\left[c^{i}\right]$ for the third dimension. The convolutional layer transformation is produced by (discretely) convolving $c^{(i)}$ filter tensors $C^{(i)}_j$, $j\in \left[d^{(i)}\right]$ of size $h\times w \times c^{(i-1)}$, where $h\times w$ is called \emph{convolution size}, with the activations in the grid structure of $V_{i-1}$, obtaining $c^{(i)}$ convolved grids of size $h^{(i)}\times w^{(i)}$, that are then concatenated to obtain the activations of $V_i$. \emph{Pooling operators} are similar operations aiming to reduce the dimensionality of the input layer, which reduce or mantain the size of the input, and not necessarily all values in the input affect each output of the layer. Given a neuron, the region of the input that affects its activations is called its \emph{receptive field}. For a thorough introduction to CNNs, we refer the reader to~\citet[Chapter~9]{deep_learning_goodfellow}. 

Fully connected feedforward networks define functions $\phi_\mathcal{N}\colon\mathbb R^{N_0}\to\mathbb R^{N_L}$, which is optimal whenever $\mathcal{X}$ and $\mathcal{Y}$ are subsets of Euclidean spaces. However, there are specific tasks, mostly classification ones, in which one deals with a finite set of points in the output space, that is, there exists $k\in\mathbb N$ such that $\mathcal{Y}=[k]$. In these cases, one selects a projection function $\pi\colon\mathbb R^{N_L}\to[k]$ that projects the outputs of the neural network $\mathcal{N}$ into the finite set of points $[k]$ and one uses $\pi\circ \phi_\mathcal{N}\colon\mathbb R^{N_0}\to[k]$ as a model. In this context, the \emph{decision region} of a label $y\in[k]$ is the set of inputs $x$ such that $\pi(\phi_\mathcal{N}(x))=y$.
To train the neural network, it is common to choose surrogate loss functions $\mathcal{L}^{\text{surr}}$ that depend directly on the output of~$\phi_\mathcal{N}$. In classification tasks, the most common configuration for a problem with $\mathcal{X}$ having $\mathbb R^d$ as an ambient space and $\mathcal{Y}=[k]$ is to choose neural networks with $N_0=d$, $N_L=k$, projection $\pi(x_1,\hdots, x_k)= \text{argmax}_{i\in[k]}\,x_i$, and surrogate loss function $\mathcal{L}^\text{surr}$ depending only on $\phi_\mathcal{N}$ like the \emph{categorical cross entropy loss} $\text{CE}$, that, in its most basic form, looks like
\begin{equation*}
\text{CE}\left(\phi_\mathcal{N}, x, y\right)= -\frac{1}{k}\text{log}\left(\frac{\text{exp}\left(\phi_\mathcal{N}(x)_y\right)}{\sum_{i=1}^k\text{exp}\left(\phi_\mathcal{N}(x)_i\right)}\right).
\end{equation*}

For neural networks with the configuration $N_L=k$ and $\pi(x_1,\hdots, x_k)=\text{argmax}_{i\in[k]}\,x_i$, there is an ambiguity when the $\text{argmax}$ is not unique, i.e., when there are at least two indices $i, j\in[k]$ such that $x_i=x_j$. In this case, the projection function must define a deterministic way to choose one of the equal-valued indices. Given a neural network $\mathcal{N}$ as the previous one, the points $x\in\mathcal{X}$ for which there exists an ambiguity is called \emph{decision boundary} of~$\mathcal{N}$. 
Formally,
\begin{equation}\label{eqn:decision_boundary_formal_definition}
\mathcal{B}\left(\mathcal{N}\right)= \Big\{x\in\text{Dom}(\phi_\mathcal{N}):\exists i,j\in[N_L]\text{ such that }\phi_\mathcal{N}(x)_i=\phi_\mathcal{N}(x)_j=\max_{\iota\in[k]}\phi_\mathcal{N}(x)_\iota\Big\}.
\end{equation}

\begin{figure}[ht]
\centering
\includegraphics[width=0.6\textwidth]{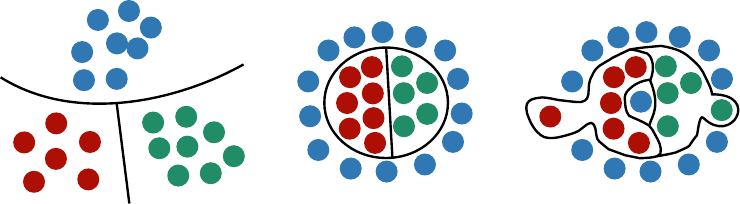}
\caption{Decision regions and boundaries for three different classification problems with three labels. The black lines represent 
decision boundaries given by FCFNNs. The different decision regions are separated by 
decision boundaries. Decision regions and boundaries give valuable information on the neural network used in each case. Classification problems are ordered from left to right by increasing \textit{complexity}. For the left and central classification problems, the decision regions and boundaries are \textit{simple}, in the sense that they seem to properly classify the inputs of the domain without visible outliers or \textit{strange} regions. However, the neural network for the central classification problem seems to have a more complex output, since the blue decision region has one hole that does not exist in the first case. This could be, for example, an indicator of the inherent difficulty of the problem. The right decision regions and boundary are more complicated. The blue decision region has two connected components and one hole, and the red and green regions have two \textit{odd} protuberances which appeared due to outliers in the training data. Although protuberances cannot be detected by the use of usual topological techniques, the extra connected component is detected by the homology of the blue decision region. See Section~\ref{scn:topo_essentials} for more details.}
\label{fig:different_decision_boundaries}
\end{figure}

Sometimes, decision regions are studied without the decision boundary; in other words,
$(\pi\circ\phi_\mathcal{N})^{-1}(y)\smallsetminus \mathcal{B}(\mathcal{N})$ is used. Figure~\ref{fig:different_decision_boundaries} shows examples of decision boundaries and regions for classification problems with three labels. 

In this context, it is usual to interpret the outputs of the function $\phi_\mathcal{N}$ as (unnormalized) probabilities, indicating the likelihood that the input belongs to one of the classes represented by the outputs. Specifically, $\phi_\mathcal{N}(x)_i$ denotes the (unnormalized) probability that $x$ belongs to the class $i\in[k]$. Therefore, $\pi$ is chosen as $\text{argmax}$ since it selects the class with the highest probability according to the output of the neural network.

For binary classification problems, i.e., $\mathcal{Y}=[2]$, the usual neural network architectures, projection functions, and decision boundaries are slightly different. On the one hand, the number of neurons in the last layer is set to one, that is, $N_L=1$. On the other hand, the usual projection function is given by $\pi(x)=2$ if $x>b$ and $\pi(x)=1$ if $x\leq b$, with the usual value of $b$ being zero. In this configuration, the decision boundary is defined as
\begin{equation}\label{eqn:decision_boundary_binary_classification}
\mathcal{B}\left(\mathcal{N}\right)= \left\{x\in\text{Dom}(\phi_\mathcal{N}): \phi_\mathcal{N}(x)=b\right\}.
\end{equation}

\subsection{Topological data analysis}\label{scn:topo_essentials}

Many tools from topological data analysis have been used within machine learning. Among them, persistent homology and Mapper graphs have so far been the most relevant in the study of deep learning methods. In this section, we provide a concise overview of both. The book by~\cite{Munkres2000} is a standard reference for general topology.
For an introduction to computational topology and to most of the concepts used in this survey, we refer the reader to the book by~\cite{edelsbrunner2022computational}.

\subsubsection{Persistent homology}
The most widely used tool from topological data analysis to analyze neural networks in this survey is persistent homology, which studies the evolution of
homology groups along an ordered family of simplicial complexes. 

An \emph{abstract simplicial complex} is a collection $K$ of non-empty finite subsets of a set $P$ such that $\{p\}$ is in $K$ for all $p\in P$ and such that $\sigma\in K$ whenever $\sigma\subseteq \sigma'$ with $\sigma'\in K$. Elements of $P$ are called \emph{vertices} and elements of $K$ are called \emph{simplices}. 
A~simplex $\sigma$ has \emph{dimension} $d$ if its cardinality is $d+1$,
and the dimension of a finite simplicial complex $K$ is the maximum dimension of its simplices. Graphs $G=(V,E)$ are $1$-dimensional simplicial complexes with $P=V$, where each edge in $E$ is a simplex $\{v,w\}$ with $v\ne w$.

Each finite simplicial complex $K$ with a prescribed order in its set of vertices has an associated family of \emph{homology groups} $H_n(K)$ for $n\in\mathbb{N}$, defined as follows. An \emph{$n$-chain} is a finite sum of $n$-dimensional simplices of~$K$ with coefficients in~$\mathbb Z$, and the \emph{boundary} of an ordered $n$-simplex $\sigma=(v_0,\dots,v_n)$ is the alternating sum $\sum_i\,(-1)^i(v_0,\dots,\hat v_i,\dots,v_n)$, where $v_i$ is omitted. An $n$-chain is an \emph{$n$-cycle} if its boundary is zero. Then $H_n(K)$ is defined as a quotient of the abelian group of $n$-cycles modulo the subgroup of $n$-boundaries. Hence, generators of $H_n(K)$ can be interpreted as $n$-dimensional ``cavities'' in~$K$.
If coefficients in a field $\mathbb F$ are used to define $n$-chains, then $H_n(K)$ becomes an $\mathbb F$-vector space. For convenience, we will work with such vector spaces in most of what follows. If no coefficient field is explicitly selected, we will assume that the one used is the field $\mathbb{F}_2$ of two elements.
The dimension of $H_n(K)$ as an $\mathbb F$-vector space is called the $n$-th \emph{Betti number} of~$K$ and denoted by $b_n(K)$. The zeroth Betti number counts the number of connected components of~$K$. For a detailed introduction to homology, we refer to~\cite{edelsbrunner2022computational}. 

An $\mathbb R$-indexed \emph{persistence module} is a family $\mathbb{V}=(V_t)_{t\in\mathbb{R}}$ of vector spaces over a field~$\mathbb F$ equipped with $\mathbb F$-linear maps $f_{s,t}\colon V_s\to V_t$ for $s\le t$, such that $f_{s,t}\circ f_{r,s}=f_{r,t}$ if $r\le s\le t$ and $f_{t,t}={\rm id}$ for all~$t$.
Given a filtration of simplicial complexes $\left(K_t\right)_{t\in\mathbb R}$ ordered by inclusion, the family
$\left(H_*(K_t)\right)_{t\in\mathbb R}$ is a persistence module, where 
$H_*(K_t)$ denotes the direct sum of $H_n(K_t)$ for all~$n\in\mathbb{N}$. The $\mathbb F$-linear maps $f_{s,t}\colon H_*(K_s)\to H_*(K_t)$ are induced by the inclusions $K_s\subseteq K_t$.
Persistence modules are discussed in detail in the book by ~\cite{chazal2016structure}.
We denote the persistence module given by the $n$-th homology $H_n$ of a filtration of simplicial complexes $\left(K_t\right)_{t\in\mathbb R}$ and fixed field $\mathbb F$ by $\mathbb V^\mathbb F_n(K)$. We drop the superscript denoting the field when it is clear from the context. 

A persistence module $\mathbb V$ is of \emph{finite type} if $V_t$ is finite-dimensional for all~$t$ and equal to zero for $t<t_0$ for some~$t_0$, and the maps $f_{s,t}$ are isomorphisms in a neighbourhood of every index value $t$ except for a finite set $\{t_1,\dots,t_k\}$ at which $f_{t_i,t}$ is an isomorphism if $t_i\le t<t_i+\varepsilon$ for some $\varepsilon>0$, yet $f_{s,t_i}$ fails to be an isomorphism if $t_i-\varepsilon<s<t_i$.

A vector $v\in V_b$ is said to be \emph{born} at a parameter value $b$ if it is not in the image of $f_{s,b}$ for any $s<b$, and a vector $v\in V_s$ \emph{dies} at a parameter value $d$ if $f_{s,d}(v)=0$ and $f_{s,t}(v)\ne 0$ for $s\le t<d$. As explained in \citep[Theorem~1.4]{chazal2016structure}, the lifetime intervals $[b,d)\subset\mathbb{R}$ of a full collection of (arbitrarily chosen) basis elements of a persistence module $\mathbb V$ of finite type represent $\mathbb V$ up to isomorphism.
The intervals $[b,d)$ form a \emph{multiset}, since they can be repeated, so every interval is given with a multiplicity. This multiset is called a \emph{barcode} of~$\mathbb V$. We assume that $d\in\bar{\mathbb{R}}$ since death parameter values are infinite in the case of vectors $v\in V_s$ for which $f_{s,t}(v)\ne 0$ for all~$t$.

\begin{figure}[htb]
\centering
\includegraphics[width=11cm]{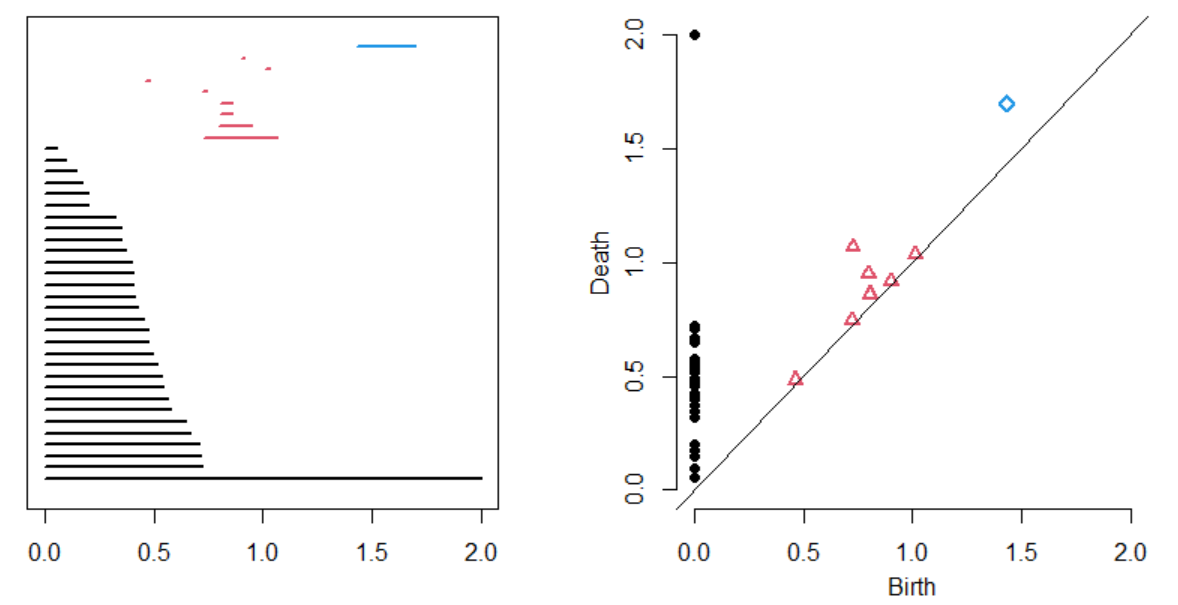}
\vspace*{-0.4cm}
\caption{Barcode and persistence diagram of a Vietoris--Rips persistence module of a point cloud with 30 points sampled from the surface of a 3D sphere of radius~$1$.}
\label{fig:barcode}
\end{figure}

Barcodes are more efficiently represented by means of \emph{persistence diagrams}, which are multisets of points in a coordinate plane with a point $(b,d)$ with $b<d$ and $d\in\bar{\mathbb{R}}$ for each interval $[b,d)$ in the barcode of a persistence module~$\mathbb V$.
We denote by $D\left(\mathbb V\right)= \left\{(b_i, d_i)\right\}_{i\in L}$ the persistence diagram of a persistence module $\mathbb V$, where $L$ is the multiset of intervals $[b_i,d_i)$ in its associated barcode; see Figure~\ref{fig:barcode}.

In the case of a filtration $(K_t)_{t\in\mathbb{R}}$ of simplicial complexes, the persistence diagram of $(H_*(K_t))_{t\in\mathbb{R}}$
describes the evolution of homology generators along thefiltration. Thus, a representative $n$-cycle $\zeta$ can be associated with each point $(b,d)$ in homological degree~$n$, so that $b$ is the birth parameter of $\zeta$ and $d$ is the value at which $\zeta$ becomes a boundary.

There are several ways to construct 
filtrations of simplicial complexes given a set of points or a weighted graph. For sets of points $P$ equipped with a symmetric function $d\colon P\times P\to\bar{\mathbb R}$ such that $d(x,x)\leq d(x,y)$ for all $x,y\in P$, called a \emph{dissimilarity}, the most frequent choice is the \emph{Vietoris--Rips filtration}, defined as
\begin{equation*}
\text{VR}_t(P,d)= \left\{\sigma\in\mathcal P(P)\smallsetminus \emptyset \,:\,\text{diam}(\sigma)\leq t\right\}.
\end{equation*}

\begin{figure}[htb]
\centering
\includegraphics[width=0.99\textwidth]{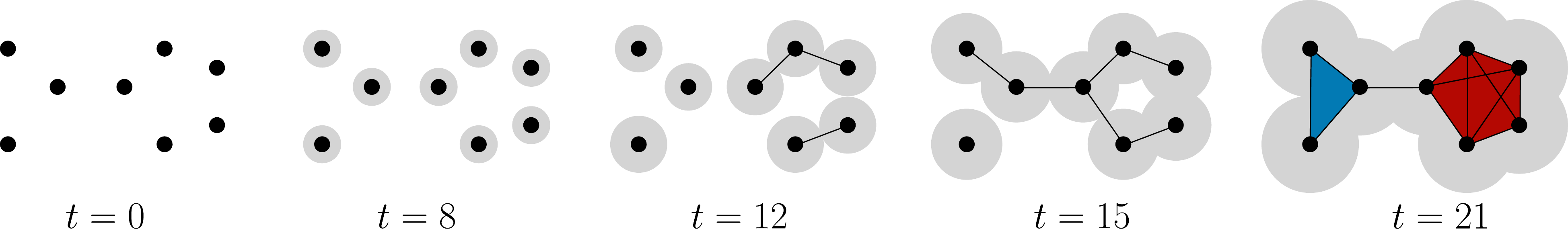}
\caption{Vietoris--Rips filtration at time values $t\in\{0, 8, 12, 15, 21\}$ for a point cloud $P$ equipped with the Euclidean distance $d=\lVert\cdot\rVert_2$ in the plane. For $t<0$, $\text{VR}_t(P,d)$ is the empty set, as points appear in the filtration at $t=0$. For $t\in\{12, 15\}$, only edges are added as there are no three vertices with pairwise distances lower than $t$. For $t=21$, one triangle and two tetrahedra are added to the filtration. Eventually, for all $t$ after a threshold, $\text{VR}_t(P,d)$ becomes a simplex of dimension equal to the number of points of $P$ minus one.}
\label{fig:vr_filtration}
\end{figure}

We refer to such pairs $(P,d)$ as \emph{point clouds}. Figure~\ref{fig:vr_filtration} shows simplicial complexes for different values of $t$ of a Vietoris--Rips filtration for a point cloud with eight points and the Euclidean distance. 
For practical purposes, there are situations in which we may want to limit the dimension of the Vietoris--Rips simplicial complexes to a maximum value $k_\text{max}$. In this case, we only take simplices up to dimension~$k_\text{max}$. We denote dimension-limited Vietoris--Rips filtrations and simplicial complexes by 
\begin{equation*}
\text{VR}^{k_\text{max}}_t(P,d) = \left\{\sigma\in \text{VR}_t(P,d)\,:\,\text{dim}(\sigma)\leq k_\text{max}\right\}.
\end{equation*}
If the point cloud $P$ is a subset of a metric space $(X,d)$, then 
another popular indexed family of simplicial complexes is the \emph{\v{C}ech filtration} 
\begin{equation*}
\check{C}_t(P, X, d)= \left\{\sigma\in\mathcal{P}(P)\smallsetminus \emptyset\,:\,\textstyle\bigcap_{x\in\sigma}\bar B(x, t/2)\neq \emptyset\right\},
\end{equation*}
where $\bar B(x,\varepsilon)= \left\{y\in X: d(x,y)\le  \varepsilon\right\}$ denotes the closed ball centered at $x$ of radius~$\varepsilon$.

Similar ideas can be applied to weighted graphs. Let $(G, w_V, w_E)$ be a weighted graph, where 
$w_V\colon V(G)\to\mathbb R$ and $w_E\colon E(G)\to\mathbb R$ are 
weight functions for the vertices and the edges of~$G$, respectively. 
By requiring that $w_V(v)\leq w_E(e)$ for all $v\in V(G)$ and all $e\in E(G)$ incident to~$v$,
the weighted graph $G$ can be treated as a point cloud.
Points correspond to the vertices of $G$ and distances between points are given by the weight functions $w_V$ and~$w_E$, where the lack of an edge between two vertices is encoded as an infinite distance. Thus the distance from a point to itself need not be zero.
Formally,
\begin{equation}
    d(v,w)= \begin{cases}
        w_V(v)&\text{ if }v=w,\\
        w_E(\{v,w\})&\text{ if }\{v,w\}\in E(G),\\
        +\infty&\text{ otherwise}.
    \end{cases}
\end{equation}
This allows one to compute a Vietoris--Rips filtration in the same way as for point clouds. We denote the Vietoris--Rips filtration of a weighted graph $(G, w_V, w_E)$ by $\text{VR}(G, w_V, w_E)$, and we denote by $\text{VR}^{k_\text{max}}(G, w_V, w_E)$ the dimension-limited version.

Sometimes, weighted graphs are provided only with weights on their edges, that is, a function $w_V$ is not provided. In such cases, there is no canonical way to define a distance between a vertex and itself. One solution is to define $w_V$ for a vertex $v\in V(G)$ as the minimum weight of all its incident edges, and another possible solution is to define $w_V$ as the global minimum weight among all edge weights. 

Vietoris--Rips and \v{C}ech filtrations add simplices with lower diameter first, that is, the lower the values of $t$, the lower the diameter of the simplices inside $\text{VR}_t$ and $\check{C}_t$. However, it is usual in the graph realm to add edges with higher weights first. 
Thus, by giving a descending order $e_1,\hdots, e_n$
of the edges by their weights, where $n=|E(G)|$, the edge $e_i$ is added at $t=i-1$ to the filtration for all $i\in[n]$. In this case, the vertices can be added either at $t=0$, or at the moment the first incident edge enters the filtration, or at $t=w_V(v)$ if the vertices are weighted by a function $w_V$ that satisfies $w_V(v)\geq w_E(e)$ for all vertices $v$ and all edges incident to~$v$. We can replicate the three approaches also using Vietoris--Rips filtrations. For the first two, we do this by defining the following dissimilarity functions:
\begin{equation*}
    d_{\downarrow}^{0}(v, w)=\begin{cases}
        0&\text{ if }v=w,\\
        i-1&\text{ if }\{v,w\}=e_i,\\
        +\infty&\text{ otherwise},
    \end{cases}\quad 
    d_{\downarrow}(v, w)=\begin{cases}
        \min\{i: v\in e_i\}&\text{ if }v=w,\\
        i-1&\text{ if }\{v,w\}=e_i,\\
        +\infty&\text{ otherwise}.
    \end{cases}
\end{equation*}
For the third one, by defining $\mathfrak{S}= E(G)\cup\left\{\left\{v\right\}: v\in V(G)\right\}$ and $w\colon \mathfrak{S}\to\mathbb R$ as  
\begin{equation*}
    w(\left\{v, w\right\})=\begin{cases}
        w_V(v)&\text{ if }\left|\left\{v, w\right\}\right|=1,\\
        w_E(\left\{v, w\right\})&\text{ otherwise},
    \end{cases}
\end{equation*}
and giving a descending order $s_1, \hdots, s_{\left|\mathfrak{S}\right|}$ of $\mathfrak{S}$ induced by $w$, we use the dissimilarity function
\begin{equation*}
    d_{\downarrow}^V(v, w) = \begin{cases}
        i-1&\text{ if }\{v,w\}=s_i,\\
        +\infty&\text{ if }\{v,w\}\not\in\mathfrak S.
    \end{cases}
\end{equation*}

For points $(b,d)$ in persistence diagrams associated with Vietoris--Rips filtrations, $b$ and $d$ are values of the corresponding dissimilarity function for some pair of points in the point cloud. This means that points of Vietoris--Rips persistence diagrams coming from functions $d_{\downarrow}^{0}$, $d_{\downarrow}$, and $d_\downarrow^V$ are tuples of indices of edges in the graph. Therefore, an alternative persistence diagram $D^w(\mathbb V_k(\text{VR}(\cdot)))$ containing the weights of the edges can be obtained simply by taking the weight values of the edges associated with the indices in the persistence diagram. That is, 
\begin{equation}
    D^w(\mathbb V_k(\text{VR}(V(G), d))) = \left\{(w_E(e_{i+1}), w_E(e_{j+1})): (i,j)\in D(\mathbb V_k(\text{VR}(V(G), d)))\right\}
\end{equation}
for $d\in\{d_{\downarrow}^{0}, d_{\downarrow}\}$, and
\begin{equation}
    D^w(\mathbb V_k(\text{VR}(V(G), d^V_\downarrow))) = \left\{(w(s_{i+1}), w(s_{j+1})): (i,j)\in D(\mathbb V_k(\text{VR}(V(G), d)))\right\},
\end{equation}
where we define $w_E(e_{\infty})=+\infty= w(s_{\infty})$. We refer to this persistence diagram as the \emph{weighted persistence diagram} of the persistence modules of the Vietoris--Rips family of filtrations for graphs induced by $d_{\downarrow}^{0}$, $d_{\downarrow}$ or $d_\downarrow^V$. This also holds for dimension-limited Vietoris--Rips filtrations. 

There are many useful filtrations that can be employed in persistent homology for finite sets of points or graphs. In this section, we have presented only some of them. For a broader perspective on point cloud and graph filtrations, we refer the reader to the works by~\cite{rieck2023expressivity}, by~\cite{edelsbrunner2022computational}, and by~\cite{Chazal2014}.

Manifolds also yield persistence diagrams computed from Morse functions. A \emph{Morse function} is a smooth function  $f\colon\mathbb M\to\mathbb R$ on a smooth manifold $M$ such that all critical points are non-degenerate, that is, the Hessian of $f$ at each critical point is non-singular. Similarly to the simplicial case, given $n\in\mathbb N$ and a field~$\mathbb F$, one can build persistence modules from the homology of sublevel sets of a Morse function at different levels $t\in\mathbb R$. Thus, the persistence module $\mathbb{V}_n^\mathbb{F}(f)$ for a Morse function $f$ on a smooth manifold is defined as
\begin{equation*}
    V_n^\mathbb{F}(f)_t= H_n(f^{-1}(-\infty, t]),
\end{equation*}
where singular homology is meant with coefficients in~$\mathbb F$. If $M$ is compact, then $\mathbb{V}_n^{\mathbb F}(f)$ is of finite type and hence a barcode and a persistence diagram can be extracted from it.

We briefly mention an extension of the usual persistence modules computed with homology that also produce persistence diagrams, known as \emph{zigzag persistence}, introduced by~\cite{zigzag_persistence}. Zigzag persistence modules are 
analogous to persistence modules, but the inclusions are not necessarily given by the order of the real numbers indexing the vector spaces. Zigzag persistence modules share a good amount of
properties with ordinary persistence modules, such as their unique representation by
persistence diagrams or their stability~\citep{stability_persistence, stability_zigzag}, under mild assumptions.

For many purposes, it is useful to have a notion of distance between persistence diagrams. The most frequently used distances between persistence diagrams are the bottleneck distance and the $q$-Wasserstein distances, for a given $q\in\mathbb N$. The \emph{bottleneck distance} between two persistence diagrams $D_1$ and $D_2$ is defined as
\begin{equation*}
    W_\infty\left(D_1,D_2\right)= \inf_{\eta: D_1^\Delta\to D_2^\Delta}\;\sup_{x\in D_1^\Delta}\lVert x-\eta(x)\rVert_\infty,
\end{equation*}
where $D_1^\Delta$ and $D_2^\Delta$ denote the persistence diagrams $D_1$ and $D_2$ supplemented with their  diagonals, that is,
$D_1^\Delta = D_1\cup \Delta$ and $D_2^\Delta = D_2\cup \Delta$, where $\Delta=\left\{(x,x):x\in\bar{\mathbb R}\right\}$,
and $\eta\colon D_1^\Delta\to D_2^\Delta$ are bijections of multisets, where points in $\Delta$ have countably infinite multiplicity. The \emph{Wasserstein distance} $W_q$ is defined as
\begin{equation*}
    W_q(D_1, D_2) = 
    \inf_{\eta: D_1^\Delta\to D_2^\Delta}\Bigg(\,\sum_{x\in D_1^\Delta}\lVert x-\eta(x)\rVert_\infty^q\Bigg)^{1/q}
\end{equation*}
although a $p$-norm $\|\cdot\|_p$ may be used instead of the supremum norm $\|\cdot\|_\infty$.
The Wasserstein distances induce norms on persistence diagrams given by the distance to the empty diagram, namely $\lVert D\rVert_q = W_q(D, \emptyset)$.

Persistence diagrams are not used directly for data analysis due to their lack of a suitable structure for statistical inference. Instead, persistence summaries are used, which are real-valued functions or vector-valued functions derived from persistence diagrams.
Examples of persistence summaries are \emph{total persistence}, which is the sum of the lifetimes of all the non-infinity points of a persistence diagram; persistence landscapes~\citep{persistence_landscapes_original}, and persistence images~\citep{persistence_images_original}, among others. 
A~\emph{landscape} $\lambda$ of a persistence diagram $D$ is a sequence of piecewise linear functions $(\lambda_1,\lambda_2, \hdots)$ defined as
\begin{equation*}
    \lambda_i(t) = \text{max}^i_{(b,d)\in D}\left\{f_{(b, d)}(t)\right\}, \quad f_{(b,d)}(t)=\max\left(0, \min\left(b+t, d-t\right)\right),  
\end{equation*}
where $\max^i$ returns the $i$-th maximum value of a multiset. 

\subsubsection{Mapper and GTDA}
\label{Mapper+GTDA}

Mapper is an algorithm, introduced by~\cite{mapper_original_paper}, that extracts visual descriptions of high-dimensional datasets in the form of simplicial complexes (usually graphs). One key concept behind Mapper is the nerve of a covering. Given a topological space $X$ and a covering $\mathcal{U}=\left\{\mathcal{U}_i\right\}_{i\in I}$ of $X$, the \emph{nerve} of $\mathcal{U}$ is defined as the simplicial complex $N\left(\mathcal{U}\right)$ whose set of vertices is $I$ and where a family $\left\{i_1,\hdots, i_k\right\}\subseteq I$ is a simplex of $N\left(\mathcal{U}\right)$ if and only if $\bigcap_{j=1}^k\mathcal{U}_{i_j}\neq \emptyset$. For good coverings (open coverings whose sets and their non-empty finite intersections are contractible), the geometric realization of $\mathcal{N}\left(\mathcal{U}\right)$ is homotopy equivalent to~$X$. Hence, 
$\mathcal{N}\left(\mathcal{U}\right)$ encodes relevant features of the shape of $X$ into a combinatorial object.

\begin{figure}[htb]
\centering
\includegraphics[width=0.55\textwidth]{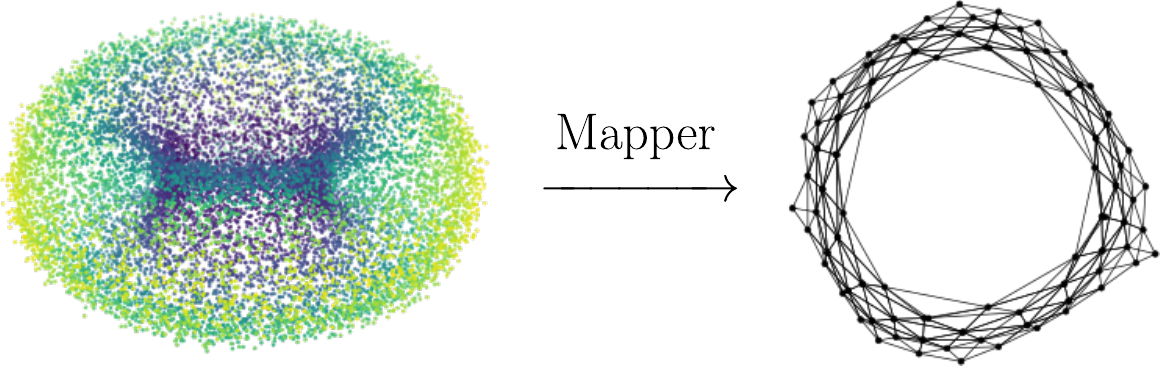}
\caption{Mapper graph generated from a noisy sample of a torus with $14{,}400$ points using Kepler Mapper~\citep{kepler_mapper_software}. The filter function used is given by the projections to the $X$ and $Y$ axes. Covers are taken with an overlap of $p=0.2$ using $100$ squares, by means of the 
standard cover implementation of the software.}
\label{fig:mapper_torus}
\end{figure}

Mapper applies the nerve construction to suitably customized coverings of datasets.
Simplicial complexes produced by Mapper are often able to reveal patterns of a latent topological space from which the dataset might have been sampled. 
The Mapper algorithm has the following ingredients as input:
\begin{enumerate*}
    \item A~dataset $\mathcal{D}$, which we assume of finite cardinality;
    \item A~filter function $f\colon \mathcal{D}\to\mathbb R^d$, where $d=1$ or $d=2$ in most use cases;
    \item A~finite cover $\mathcal{U}=\left\{\mathcal{U}_i\right\}_{i\in I}$ of $\text{Im}(f)\subset\mathbb R^d$ consisting of overlapping open sets;
    \item A~clustering algorithm, such as \texttt{DBSCAN}~\citep{dbscan_clustering}.
\end{enumerate*}
The method is described in Algorithm~\ref{alg:mapper_basic} in the Appendix.

Mapper is most often applied using filter functions $f\colon\mathcal{D}\to\mathbb R$ 
and choosing covers $\mathcal{U}$ consisting of intervals
with a fixed percentage of overlap~$p$. With these choices and using clustering algorithms that generate disjoint clusters, the simplicial complexes produced by Mapper are graphs. Mapper graphs are closely related with Reeb graphs
\citep[Chapter~VI]{reeb_graphs_original,edelsbrunner2022computational}.
For an advanced analysis of the Mapper algorithm for filter functions $f\colon\mathcal{D}\to\mathbb R$ and its connection with Reeb graphs, we refer the reader to the work by~\cite{structure_and_stability_mapper}.

It was observed by~\cite{gtda_for_decision_regions} that, 
in some experiments studying the output space of neural networks, Mapper produced too many tiny components that did not carry useful information for the experiments. 
For this reason, they introduced a new algorithm called \emph{graph-based topological data analysis} (GTDA). The GTDA algorithm builds on the Mapper algorithm to construct Reeb networks from graph inputs instead of point clouds. Reeb networks generalize Reeb graphs. Each vertex in a Reeb network built using GTDA is associated with a subgraph of the original graph. Given an input graph~$G$, GTDA uses filter functions $f\colon V(G)\to\mathbb R^d$ to build Reeb network vertices using a recursive splitting strategy based on filter function values. After initial vertices are generated, the smallest ones are merged into other vertices and edges are added. Vertices within connected components with non-empty intersection are connected, and other edges are added to promote connectivity of the output. The GTDA algorithm is described  in~\citet[Algorithm~1]{gtda_for_decision_regions}.

\subsection{Applications of topological data analysis in deep learning}\label{sec:applications_of_tda_in_dl}
Some of the methods for analyzing neural networks in this survey have been applied to solve corresponding deep learning problems. In this section, we summarize and contextualize these deep learning problems.

\subsubsection{Regularization}\label{scn:regularization_application}

In machine learning, regularization techniques are \textit{algorithmic tweaks} intended to reward models of lower complexity~\citep{understanding_nn_still_requires_generalization}. In the context of deep learning,  regularization techniques aim to reduce model overfitting, that is, to avoid neural networks that excel in minimizing empirical risk $\widehat{\mathcal{R}}_{\mathcal{D}_\text{train}}$ but generalize poorly to a low real risk $\mathcal{R}$.

There are many ways in which regularization is performed in deep learning. Two of the most important ones are early stopping and the use of regularization terms. 

In early stopping, training is stopped whenever a quality measure $\mathcal{Q}$, usually the empirical risk, evaluated on a set $\mathcal{D}_\text{val}$ different from the training dataset $\mathcal{D}_\text{train}$ stops improving. Defining the quality measure and the moment at which we consider that this measure has stopped improving is not an easy task, and it depends on the properties of the neural network and the training procedure.  Some (non-topological) tips and tricks to tune early stopping can be found in~\cite{Prechelt1998}.

Regularization terms are (almost everywhere) differentiable functions $\mathcal{T}\colon\Theta\to\mathbb R$ depending on the parameters of the neural network architecture being trained. They are minimized next to the empirical risk $\widehat{\mathcal{R}}_{\mathcal{D}_\text{train}}(\theta)$, i.e., making the training algorithm minimize the quantity
\begin{equation*}
\widehat{\mathcal{R}}_{\mathcal{D}_\text{train}}(\theta) + \alpha \mathcal{T}(\theta),
\end{equation*}
with respect to the parameters $\theta$, where $\alpha\in\mathbb{R}_{> 0}$ is called \emph{learning rate}, and controls the influence of the regularization term over the empirical risk.
\subsubsection{Pruning of neural networks}
Modern neural networks typically consist of highly complex architectures with a large number of parameters. Working with such models usually requires large amounts of computational resources that may not be easily available or sustainable in the long term. One of the approaches to solve this problem is neural network pruning. Neural network pruning involves systematically removing parameters from an existing network so that the resulting network after this process conserves as much performance as possible, but with a drastically reduced number of parameters~\citep{neural_network_pruning}.

\subsubsection{Detection of adversarial, out-of-distribution, and shifted examples}
Adversarial, out-of-distribution, and shifted examples are specific values in the input space of a neural network that have special properties with respect to the underlying distribution of the input data or the way the neural network processes them. Given a machine learning model, adversarial examples are inputs that differ slightly from correctly classified inputs but are misclassified by the model. Differences between adversarial examples and properly classified regular examples are often imperceptible to human senses~\citep{goodfellow2015explaining}. On the other hand, out-of-distribution examples are valid input values for a machine learning model that have not been sampled from the real data distribution. Similarly, shifted examples are examples that deviate from the real data distribution but come from samples from it, resembling the nature of adversarial examples~\citep{topological_uncertainty}. 

Detecting input types is an important problem in deep learning. On one hand, it allows further study of the robustness and performance of a given neural network. On the other hand, it allows to detect attacks to machine learning models, both in training and in inference.

\subsubsection{Detection of trojaned networks}\label{scn:trojaned_neural_networks}
In a Trojan attack on a machine learning model
\citep{topo_detection_trojaned},
attackers create \textit{trojaned} examples, that are normal training examples to which some extra information, using specific patterns, is added. This extra information is used by attackers to generate specific outputs in the inputs containing these specific patterns, allowing attackers to bypass the regular outputs of the model during inference. To be able to generate these \textit{unwanted} outputs, attackers inject trojaned examples into the original training data set without the knowledge of the legitimate owner of the model before the model is trained with the objective that the network recognizes the pattern and acts according to the attackers' intention. Detecting neural networks trained with trojaned examples is vital to avoid security breaches in machine learning, especially in critical-context applications such as medical systems or security systems, among others.

\subsubsection{Model selection}\label{scn:model_selection}
Over the last years, there has been an explosion on the quantity and quality of models developed by deep learning researchers and practitioners. Most of these models have a large number of parameters and are hard and expensive to train. However, once trained successfully, they generalize well to many situations. For this reason, a reasonable practice in nowadays applications is, instead of training from zero new models, taking pre-trained models and adapting them to the particular problem the user is trying to solve. However, selecting the pre-trained model that best fits the user's requirements from a \textit{bank} of models is not a trivial task. For starters, each bank of models has its own models, trained in different datasets with different algorithms and different hardware. Small variations of the parameters in the same architecture can induce important differences during inference, so selecting only by the architecture is not usually the best option. Additionally, due to the growing number of options available, testing all of them may be an unfeasible process for practical applications. For this reason, a good model selection algorithm that takes into account many properties of the models is fundamental. 

\subsubsection{Prediction of accuracy}
In classification tasks, a usual loss function is given by
\begin{equation}
    \mathcal{L}(\mathcal{N}, x, y) = \mathbbm{1}_{\{(x,y)\,:\,x\neq y\}}\left(\mathcal{N}(x), y\right).
\end{equation}
Assuming that the values of $\mathcal{Y}$ are fully determined by a function $f\colon\mathcal{X}\to\mathcal{Y}$ in the data distribution, 
the real risk $\mathcal{R}(\mathcal{N})$ for this loss function is the expectation that the neural network $\mathcal{N}$ does not match the function $f$ for an input value~$x$, that is, misclassifies~$x$. If instead we consider the value $1-\mathcal{R}(\mathcal{N})$, this is the expectation that the neural network correctly classifies an input $x$, i.e., the probability of $\mathcal{N}$ matching $f$. We define this value by accuracy and denote it by $\text{Acc}(\mathcal{N})= 1-\mathcal{R}(\mathcal{N})$. The higher the accuracy, the better the model according to the risk of $\mathcal{L}$. The problem with $\text{Acc}(\mathcal{N})$ is that we cannot compute it, so we approximate it by the empirical accuracy of $\mathcal{N}$ in a sample $\mathcal{D}$ different from the training dataset $\mathcal{D}_\text{train}$, usually called test dataset and denoted by $\mathcal{D}_\text{test}$. Therefore, this empirical accuracy in $\mathcal{D}$ can be computed as
\begin{equation}
   \widehat{\text{Acc}}_\mathcal{D}(\mathcal{N})= \frac{1}{\left|\mathcal{D}\right|}\sum_{(x,y)\in\mathcal{D}}\mathbbm{1}_{\{(x,y)\,:\,x = y\}}\left(\mathcal{N}(x), y\right).
\end{equation}
The empirical accuracy on a test dataset $\mathcal{D}_{\text{test}}$ is one of the most used metrics to assess the generalization and performance of a trained neural network and is equivalent to studying a numerical approximation of the risk function and thus, the generalization capacity of the neural network. Due to the hardness of proving tight bounds on the real risk $\mathcal{R}$ of a neural network given $\mathcal{L}$, many works focus on predicting the empirical accuracy $\widehat{\text{Acc}}_{\mathcal{D}_\text{test}}(\mathcal{N})$ or a variation called empirical generalization gap, given as the difference between the accuracy in a test dataset and the accuracy in the training dataset, i.e., 
\begin{equation*}
    \widehat{\text{Gap}}(\mathcal{N})=\widehat{\text{Acc}}_{\mathcal{D}_\text{test}}(\mathcal{N}) - \widehat{\text{Acc}}_{\mathcal{D}_\text{train}}(\mathcal{N}).
\end{equation*}
An analysis of methods that try to predict this last measure can be found in~\cite{pmlr-v133-jiang21a}. Also, a large-scale study of generalization bounds and measures of generalization in deep learning can be found in~\cite{Jiang*2020Fantastic}.
\subsubsection{Quality assessment of generative models}\label{scn:quality_assessment_generative_models}
Neural network quality measurements in classification tasks are more straightforward to define than in generative tasks. In general, there are many interesting factors that can be evaluated to compare the quality of two generative neural networks, and thus deciding which is better is a hard challenge. For example, one of these factors could be the similarity between the supports of the original data distribution $\mathbb P_X$
and the distribution induced by the generation function. 

\section{Analysis of neural networks using topological data analysis}\label{scn:analysis_dnn_using_tda}

In this section, we discuss the articles that use topological data analysis to study neural networks. This section is divided into the four categories described in the Introduction. Table~\ref{tab:summary_articles} in the appendix contains a list of all articles included in the survey. For each paper, we specify a one-line summary of it, its categories, and the applications from Section~\ref{sec:applications_of_tda_in_dl} explored in the article.

\subsection{Structure of a neural network}\label{scn:structure_of_the_neural_network}

Recall that a fully connected feedforward neural network $\mathcal{N}$ is specified by an architecture $a=(N, \varphi)$ and a set of parameters $\theta$ depending on the architecture $a$. At the same time, recall that the architecture $a$ defines a directed graph $G(\mathcal{N})$ that characterizes the computations of the neural network function, where each vertex is associated with some value during the evaluation of $\phi_\mathcal{N}$. The structure of this graph is relevant for studying the generalization capacity and expressiveness of neural networks. Note that edge directions are important in the neural network graph $G(\mathcal{N})$ because they characterize how computations are performed on the data. For this reason, it is preferable to avoid analyzing $G(\mathcal{N})$ forgetting its directions with the general topological tools presented in Section~\ref{scn:topo_essentials}. 

\cite{path_homologies_of_deep_feedforward_networks} studied and compared two special homology groups designed for directed graphs in the context of FCFNNs. More specifically, the ranks of these two variants of homology groups are computed. These groups are the path homology~\citep{grigoryan2014homotopy} and the directed flag complex (DFC) homology~\citep{directed_flag_complex} groups, denoted by $\text{PathHom}_k(G(\mathcal{N}))$ and $\text{DFCHom}_k(G(\mathcal{N}))$, respectively, where $k$ denotes the dimension of the homology groups as in the simplicial case. In particular, \citeauthor{path_homologies_of_deep_feedforward_networks}\ showed that, given a neural network $\mathcal{N}$ with architecture $a=(N,\varphi)$, where $N=(N_0,\hdots, N_L)$, and given a dimension $k\in\mathbb N$, the following equalities hold:
\begin{equation*}
    \begin{split}
\text{rank}\left(\text{PathHom}_k(G(\mathcal{N}))\right) &= \delta_{L-1, k}\prod_{i=0}^L(N_i-1),\\        \text{rank}\left(\text{DFCHom}_k(G(\mathcal{N}))\right) &= \text{rank}\left(H_k(G(\mathcal{N}))\right),
    \end{split}
\end{equation*}
where $\text{rank}\left(H_k(G(\mathcal{N}))\right)$ is the rank of the
$k$-dimensional simplicial homology group of $G(\mathcal{N})$ seen as an undirected graph. It can be further shown that \begin{equation*}
\text{rank}\left(H_k(G(\mathcal{N}))\right) = \begin{cases}
            1 &\text{if }k=0,\\
            1-\left|V(G(\mathcal{N}))\right|+\left|E(G(\mathcal{N}))\right| &\text{if }k=1,\\
            0&\text{if }k\geq 2.
        \end{cases}
\end{equation*}

The equality between the DFC-homology ranks and the standard simplicial homology groups for neural network graphs implies that $\text{DFCHom}$ does not capture the information carried out by the directions of neural networks. Note that the ranks for both families of homology groups can be directly computed from the number of neurons and edges contained by the network, and do not reflect most of the structural properties of the neural network graph. For example, path homologies are invariant to layer permutations, although the order of the layers is a key factor in achieving good neural network performance. On the other hand, DFC-homology forgets all the structural information except for the number of vertices and edges, which many neural network configurations with different performances share. 

Path homology theory for neural networks has an important drawback, namely that it forgets the weights of neural networks, which are crucial in the performance of neural networks. However, this can be remedied by studying a \textit{persistent} version of it, called \emph{persistent path homology}~\citep{persistent_path_homology}. To the best of our knowledge, there is no work using persistent path homology to analyze neural networks, and it could be a notable future work for the interested reader. 

\subsection{Input and output spaces}\label{scn:input_and_output_spaces}
In this section, we analyze how TDA has been used to analyze decision regions, boundaries, decompositions of the input space, and input and output spaces of generative and non-generative models.

\category{General output space and decision regions}

\noindent Let $f\colon \mathcal X\to [k]$, $k\in\mathbb N_{\geq 2}$ be the unknown function that defines a classification problem as in Section~\ref{scn:deep_learning_fundamentals} with data distribution $\mathbb{P}_{(X,Y)}$. A simple remark is that if we train a perfect neural network $\mathcal{N}$ with projection function $\pi$, that is, $\pi\circ\mathcal{\phi}_\mathcal{N}=f$, then the homology groups of the decision regions of each label $i\in[k]$ must coincide with the homology groups of the space of data with the same label for all dimensions, that is, $H_k\left((\pi\circ\phi_\mathcal{N})^{-1}(i)\right)=H_k\left(f^{-1}(i)\right)$ for all $k\in\mathbb N$ and $i\in [k]$. The previous equality suggests that the topology of the decision regions is relevant for understanding the generalization of trained neural networks with respect to the data. 

One of the first articles in this survey that studied decision regions using topological data analysis was published by~\cite{on_the_complexity_of_neural_network_classifiers_a_comparison_between_shallow_and_deep_architectures}. In it, Betti numbers of the decision regions of one of the labels of a binary classification problem are analyzed and theoretically bounded for neural networks with Pfaffian activations~\citep{pfaffian_functions_original}. A~difficulty in the previous paper is that bounds of Betti numbers are not tight and cannot be used easily in practical scenarios. An experimental approach is taken by~\cite{on_characterizing}, who empirically study such Betti numbers and compare them with the Betti numbers of the real region induced by the selected label, that is, $f^{-1}(i)$, where $i\in[2]$ is the label studied.

On a more general setting, \cite{gtda_for_decision_regions} proposed the GTDA algorithm of Section~\ref{Mapper+GTDA} as an improved version of Mapper to study the output space of general neural network functions. They studied Reeb networks built using GTDA with input graphs induced from the output vectors $(\phi_\mathcal{N}^{(L)}(x)_1,\hdots, \phi_\mathcal{N}^{(L)}(x)_{N_L})$ of neural network functions given a dataset $\mathcal{D}$ and filter functions given by the same output values plus, possibly, some extra values depending on the dataset or the outputs. To give output values a graph structure, they either used previously-known binary relationships between the data or binary relationships given by nearest neighbors algorithms as edges.

\citeauthor{gtda_for_decision_regions}\ were able to gain useful information from GTDA graphs that was not found using Mapper or other alternatives about output values of the Enformer model~\citep{ensformer_model} used as inputs to predict harmful mutations of a human gene. Also, GTDA graphs were used to perform automatic error estimation for machine learning prediction tasks. A numerical error estimation score for a sample $x$ was calculated by comparing the predicted label of $x$ with ground truth labels of training and validation samples whose prediction vectors were surrounding the prediction vector of $x$ in the union of subgraphs of the Reeb network vertices plus the extra edges added in the GTDA algorithm. The error estimation score was shown to successfully correct erroneous predictions in binary classification tasks. Also, it detected regions of the output space with many misclassifications, that allowed them to determine the source of error in those regions for the previous classification problem dealing with the Enformer architecture. Similar experiments were performed on other network architectures and datasets such as ResNet-50 and Imagenette~\citep{imagenette}, showing that GTDA offers unique insights about the output space that cannot be recovered with other methods such as Mapper. 

\category{Decision boundaries of a single network}

\noindent When using classification neural networks with $N_L=k$ and $\pi(x_1,\hdots, x_k)=\text{argmax}_{i\in[k]}\,x_i$, decision boundaries $\mathcal{B}$, defined in Equation~\eqref{eqn:decision_boundary_formal_definition}, become relevant as they determine the decision regions of the neural network and the spaces in which neural network decisions have more than one valid output. \cite{tda_of_decision_boundaries_with_application_to_model_selection} propose a modified version of the \v{C}ech simplicial complexes such that, given a large enough sample of points in the decision region of a binary classification neural network, can reconstruct a space that is homotopy equivalent to the real decision boundary induced by the neural network under some mild assumptions about the decision boundary. However, \v{C}ech complexes are too costly to compute in practical scenarios. For this reason, they propose two computationally feasible variants of the Vietoris--Rips filtrations and complexes that aim to recover the Betti numbers of the decision boundaries of neural networks from a sample of their points using the persistence diagrams induced by these new filtrations. The new filtrations $\left(\text{P-LVR}_t(\mathcal{D})\right)_{t\in\mathbb R}$ and $\left(\text{LS-LVR}_t(\mathcal{D})\right)_{t\in\mathbb R}$, called plain-LVR and locally scaled LVR, respectively, are constructed from a sample $\mathcal{D}=\left\{(x_1,y_1),\hdots,(x_m, y_m)\right\}$ containing elements from both classes. Specifically, given $\bullet \in \left\{\text{P-LVR}, \text{LS-LVR}\right\}$, the simplicial complexes of these filtrations are
\begin{equation*}
   \bullet_t(\mathcal{D}) = \{\sigma\subseteq \{x_i\}_{i=1}^m: \{\sigma_i, \sigma_j\}\in G_t^{\bullet}(\mathcal{D})\text{ for all } \{\sigma_i,\sigma_j\}\subseteq \sigma\}, 
\end{equation*}
where $G_t^{\bullet}(\mathcal{D})$ is the simplicial complex of dimension $1$ consisting of the vertices and edges of a bipartite graph $G^{\bullet}_t=(V_1, V_2, E_t)$ with $V_j=\left\{x_i: y_i=j\right\}$ for $j\in[2]$ satisfying $\{v_1, v_2\}\in E_t$ if and only if $d_\bullet(v_1, v_2)\leq t$, plus a set of new edges joining the vertices connected by paths of length two in $G^{\bullet}_t$. The dissimilarities $d_\bullet$ are given by
\begin{equation*}
    d_\text{P-LVR}(v_1, v_2)= \lVert v_1 - v_2\rVert_2, \qquad d_\text{LS-LVR}(v_1, v_2) = \frac{\lVert v_1 - v_2\rVert_2}{\sqrt{\rho_1\rho_2}},
\end{equation*}
where $\rho_i$, $i\in[2]$ is the local scale of $v_i$, that is, the radius of the smallest sphere centered at $v_i$ that encloses at least $k$ points from the opposite class, where $k$ is a fixed parameter.

A~problem with the previous constructions is that they may require very large samples to robustly recover the topology of the decision boundaries. Obtaining such samples can sometimes be infeasible, especially when the cost of evaluating the value of $y\in\mathcal{Y}$ by the unknown function $f$ given $x\in\mathcal{X}$ is high. \cite{finding_the_homology_of_decision_boundaries_with_active_learning} propose a method based on active learning~\citep{settles2009active} to sample a small but meaningful set to recover the homology of the decision boundaries using the labeled \v{C}ech and Vietoris--Rips complexes introduced in the article. Information extracted from persistence diagrams is used in both articles to perform model selection for basic datasets including Banknote~\citep{banknote_dataset}, CIFAR-10~\citep{cifar-10}, MNIST~\citep{MNIST_article}, and Fashion-MNIST~\citep{fashionMNIST}, and basic machine and deep learning models.

Decision boundaries are directly related to the performance of neural networks. Given a binary classification task and a neural network $\mathcal{N}$ with $N_L=1$ and projection function $\pi(x)=2$ if $x> 0$ and $x \leq 0$, recall that the decision boundary of $\mathcal{N}$ is the set $\phi_\mathcal{N}^{-1}(0)$, as in Equation~\eqref{eqn:decision_boundary_binary_classification} for $b=0$. Let $x\in\text{Dom}(\phi_\mathcal{N})$. For this kind of neural network, the higher the absolute value of $\phi_\mathcal{N}(x)$, the more likely it is that the same input, slightly perturbed, is classified with the same label as~$x$, that is, $\pi(\phi_\mathcal{N}(x))=\pi(\phi_\mathcal{N}(x+\varepsilon))$, for a small~$\varepsilon$. In other words, the higher the absolute value of $\phi_\mathcal{N}$  for a given example $x$, the more robust the prediction of $\mathcal{N}$ for $x$. 

\cite{a_topological_regularizer_for_classifiers_via_persistent-homology} propose a regularization term to remove \textit{weak} connected components from the decision boundary, that is, connected components enclosing decision regions such that the predictions for their points are not robust. This regularization term for binary classification tasks can be computed from zero-dimensional zigzag persistence diagrams induced by $\phi_\mathcal{N}$ whenever the neural network function $\phi_\mathcal{N}$ is Morse, which occurs in several neural network architectures~\citep{Kurochkin2021}, and $\mathcal{X}$ is a hypercube. This regularization term was the first to use the differentiability properties of persistent homology, studied and introduced in parallel by~\cite{a_framework_for_differential_calculus_on_persistence_barcodes} and by~\cite{optimizing_persistent_homology_based_functions}. The capacity of the term was studied for a simple kernel logistic regression, that was compared to other non-deep learning models such as  KNN, logistic regression, or SVM, among others, trained with $L^1$ and $L^2$ regularization terms. The experiments were performed in synthetic and real datasets, where the real datasets came from the UCI dataset bank~\citep{uci_machine_learning_repo} and from two biomedical datasets~\citep{Yuan2014-gv, fMRI_dataset_weak_topology}. In most cases, the model trained with the topological regularizer outperformed the rest of the models. However, no deep learning models were involved, and a further experimentation is needed to see the efficacy of this approach for neural networks. 

\category{Decision boundaries of a hypothesis set}

\noindent Previous articles have studied decision regions and boundaries for specific instances of neural networks, that is, pairs consisting of an architecture $a$ and a particular set of parameters~$\theta$. \cite{on_the_topological_expressive_power_of_neural_networks} study the topology of all possible decision boundaries given by a hypothesis set $\mathcal{F}_{a,\Theta}$ of a given fixed architecture~$a$, and a given fixed set of possible parameters~$\Theta$. The objective of this study is to associate a \textit{topological diversity measure} to a particular hypothesis set $\mathcal{F}_{a,\Theta}$ that characterizes the diversity in the topology of the different decision boundaries induced by the hypothesis set. To do this, a sample is first taken with a large number of possible parameters $\left\{\theta_1,\hdots,\theta_m\right\}\subseteq\Theta$. Then, for each parameter $\theta_i$, $i\in[m]$, the zero and one-dimensional Vietoris--Rips persistence diagrams $D(\mathbb V_k(\text{VR}(P_i, \lVert \cdot\rVert_2)))$ of a sample of points $P_i$ from an approximation of the decision boundary of the neural network with architecture $a$ and parameters $\theta_i$ are computed. Finally, topological diversity measures of the hypothesis set are given by the spread of the metric spaces~\citep{spread_metric_spaces} whose points are the computed persistence diagrams for a fixed dimension and whose distance is a $q$-Wasserstein distance for a fixed $q\in\mathbb N$. To approximate spreads, which are computationally expensive to compute, Petri and Leit{\~a}o saw that the averages of the $1$-norms of the previously computed persistence diagrams, which are easier to compute, correlate with their spreads and thus also characterize the topological diversity of the hypothesis set. 

\category{Input space decomposition into convex polyhedra} 

\noindent Neural networks with a ReLU activation function $\varphi(x)=\max(0, x)$ determine a decomposition of the input space into convex polyhedra that assigns to each polyhedron in the decomposition a unique binary vector in such a way that two polyhedra share a facet if and only if their associated binary vectors differ exactly in one component. \cite{relu_neural_networks_polyhedral_decompositions} use this polyhedral decomposition of the input space to infer the topology of manifolds embedded in the domain of a given ReLU neural network from a sample of their points using persistence diagrams. To compute persistence diagrams given the sample $S=\{x_i\}_{i=1}^m$ of the manifold $\mathbb M$, the proposed procedure computes the set of unique binary vectors $\mathfrak B(S)= \left\{b(x_i)\right\}_{i=1}^m$ associated with the points of the sample, where $b(x_i)$ is the binary vector associated with the convex polyhedra in which $x_i$ lies, and then computes the Vietoris--Rips persistence diagrams $D(\mathbb V_k(\text{VR}(\mathfrak B(S), h)))$ of $\mathfrak B(S)$ using the Hamming distance~$h$, which counts the number of different bits in the two binary strings to compare. Although this is not directly related to neural network analysis, we firmly believe that this opens a way to study the structure and topology of the unknown data distribution for a specific learning problem by applying this method to large enough data samples.

\cite{masden2022algorithmic} used polyhedral decompositions to study the topology of decision boundaries of small, non-trained, ReLU neural networks by attaching a polyhedral complex structure to the decomposition. Polyhedral complexes are similar to simplicial complexes, in that they are sets of polyhedra such that every face of a polyhedron of the complex is also in the complex, and the intersection of any two polyhedra is either empty or a face of both. \citeauthor{masden2022algorithmic}\ found that polyhedral complexes derived from neural networks under some technical assumptions possess a dual structure known as \emph{signed sequence cubical complexes}, and provided a polynomial time algorithm with respect to the number of non-input neurons to compute both complexes, yet exponential with respect to the input dimension. The decision boundary of the neural network corresponds to specific polyhedra, which in turn have dual representations in the signed sequence cubical complexes, from which homology groups for the decision boundary can be recovered under certain assumptions, such as the need of performing a one-point compactification of the decision boundary.  

Differences were found in the structure of decision boundaries in terms of their neural network layer structures. In networks with a single hidden layer, the topology of decision boundaries exhibited remarkable consistency across various parameter ranges, including modifications to input data dimension and the number of hidden neurons. In contrast, networks with two hidden layers displayed more variability in the topological distribution of their decision boundaries.

\category{Generative neural networks}

\noindent TDA has also been used in generative deep learning. One of the most well-known models for generative deep learning is the generative adversarial network (GAN)~\citep{goodfellow2014generative}.  Roughly, generative adversarial networks try to generate synthetic samples from a \textit{real} data space $\mathcal{X}\subseteq \mathbb R^d$ distributed by an unknown function $\mathbb P_X$. To do this, generative adversarial models are composed of two neural networks, the generator $\mathcal{G}$, and the discriminator $\mathcal{S}$, with functions $\phi_\mathcal{G}\colon\mathcal{Z}\to\mathbb{R}^d$ and $\phi_\mathcal{S}\colon \mathbb R^d\to \{0, 1\}$, respectively, where $\mathcal{Z}$ is a space distributed by a known distribution $\mathbb P_Z$. We refer to $\mathcal Z$ as the noise space of the generative adversarial network. The generator is the neural network whose objective is to generate synthetic samples from samples from the noise space. On the other hand, the discriminator is the network that, given a sample, generated by $\mathcal{G}$ or sampled from the real data space $\mathcal{X}$, tries to distinguish between the two. The training of the generative adversarial network is performed by training the generator and the discriminator in an adversarial way, that is, the generator tries to fool the discriminator and the discriminator tries to distinguish between the synthetic and the real samples. 

One of the problems with generative models is the evaluation procedure. Quantitative evaluation of the quality of a model is not straightforward. To this end, \cite{geometry_score_comparing_generative_adversarial_networks} propose the \textit{Geometry Score}, a persistent homology-based quality metric for generative adversarial networks. This metric is based on the following three assumptions:
\begin{enumerate*}
    \item The higher the quality of the generator, the more similar are the spaces $\text{supp}(\mathbb P_X)$ and $\phi_\mathcal{G}(\text{supp}(\mathbb P_Z))$;
    \item $\text{supp}(\mathbb P_X)$ is concentrated in a manifold $\mathcal{M}_X$ and $\phi_\mathcal{G}(\text{supp}(\mathbb P_Z))$ is a manifold;
    \item The higher the similarity between the spaces $\mathcal{M}_X$ and $\phi_\mathcal{G}(\text{supp}(\mathbb P_Z))$, the higher the similarity between their topologies.
\end{enumerate*}
Note that the first part of the second assumption is the manifold hypothesis, and it may not hold in general scenarios~\citep{topological_singularity_detection_at_multiple_scales}. However, this detail is irrelevant for persistent homology of point clouds, as it works for samples of spaces with or without a topology. 

To compare both spaces $\mathcal{M}_X$ and $\phi_\mathcal{G}(\text{supp}(\mathbb P_Z))$, \citeauthor{geometry_score_comparing_generative_adversarial_networks}\ compute persistence modules of witness filtrations~\citep{witness_complex} for samples from $\mathcal{X}$ and $\phi_\mathcal{G}(\text{supp}(\mathbb P_Z))$, asuming that the samples taken from $\mathcal{X}$ are also contained in $\mathcal{M}_X$. Witness filtrations have the advantage that they are usually efficient even for big point clouds, although they may be more challenging to use effectively. Given a point cloud $(P,d)$, a subset of \textit{landmarks} $L\subseteq P$, and $\alpha \geq 0$, the $\alpha$-witness complex of $(P,d)$ with landmarks $L$ is the simplicial complex
\begin{equation}
    W(P, L, \alpha, d)= \left\{\sigma\subseteq L: \exists w\in X\;\forall l\in\sigma\;\forall l'\in L\smallsetminus\sigma,\;d(w,l)^2\leq d(w, l')^2+\alpha\right\}.
\end{equation}

Choosing $\alpha_\text{max} > 0$, we have that $W_{\alpha_\text{max}}= \left(W(P, L, \alpha, d)\right)_{\alpha\in[0,\, \alpha_\text{max}]}$ is a filtration of simplicial complexes that induces a persistence module $\mathbb V_k(W_{\alpha_\text{max}})$ where, for $t>\alpha_\text{max}$, all the vector spaces forming the persistence module are equal to the one corresponding to the value $t=\alpha_\text{max}$. 

To measure the difference between the two persistence modules induced by the spaces $\mathcal{M}_X$ and $\phi_\mathcal{G}(\text{supp}(\mathbb P_Z))$, \citeauthor{geometry_score_comparing_generative_adversarial_networks}\  define the relative living times $\text{RLT}$ of $i$-features for $k$-dimensional homology, given by
\begin{equation}
    \text{RLT}(i,k,P,L,d)=\frac{\mu\left(\left\{\alpha\in [0, \alpha_\text{max}]:\beta_k(W(P, L, \alpha, d))=i\right\}\right)}{\alpha_\text{max}}.
\end{equation}
Note that the persistence modules, and thus the relative living times, are dependent on a choice of landmarks $L$. To minimize the effect of this dependency, \citeauthor{geometry_score_comparing_generative_adversarial_networks}\ set a fixed size of the set of landmarks $L$ and compute the mean relative living times, given by the expectation of the relative living times over all possible choices of landmarks $L$ of the fixed size, i.e., $\text{MRLT}(i, k, P,d)=\mathbb E_L[\text{RLT}(i,k,P,L,d)]$. Finally, the Geometry Score of a generative adversarial network $\mathcal{G}$ is given by the squared differences of all the mean relative living times for all the $i$ for the $1$-dimensional persistence modules generated by samples $P$ and $P'$ equipped with the same dissimilarity function $d$ from $\mathcal{X}$ and $\phi_\mathcal{G}(\text{supp}(\mathbb P_Z))$, respectively, that is,
\begin{equation}
    \sum_{i=0}^\infty \left(\text{MLRT}(i, 1, P, d) - \text{MLRT}(i, 1, P', d)\right)^2.
\end{equation}
The validity of the Geometry Score as a quality metric was proved in several experiments. For example, it was ratified that the WGAN-GP~\citep{wgan_gp_model} is better than the WGAN model~\citep{wgan_model}, a well-known result, for the MNIST dataset. Additionally, the Geometry Score was able to distinguish between good and bad generative models derived from the DCGAN architecture~\citep{dcgan_model} trained on the CelebA dataset~\citep{liu2015faceattributes}.

In a very similar fashion, but with a simpler approach, \cite{charlier2019pho} propose to compare generative models, this time not necessarily restricted to GAN models, by directly comparing the zero and one-dimensional persistence diagrams induced by Vietoris--Rips filtrations of samples taken from the real data distribution and from the generative model using the bottleneck distance. In this case, the experiments compared four different generative models in the credit card fraud detection dataset~\citep{credit_dataset}. These models were WGAN and WGAN-GP together with WAE~\citep{tolstikhin2018wasserstein} and VAE~\citep{vae_model}. In this case too, WGAN-GP was shown to produce better results than the other models.

The two aforementioned approaches evaluating the quality of generative models were based on comparing the topology of real and synthetic data by means of persistent homology. However, they are not the usual choices in the generative deep learning community, where scores such as the Fr\'echet inception distance~\citep{fid_metric} or a numerical approximation of precision and recall~\citep{precision_and_recall_metric} are currently preferred to evaluate generative models. \cite{kim2023toppr} propose an alternative way to approximate precision and recall scores using an approximation of the support of the data distributions based on preimages of kernel density estimators (KDE) that preserve the topology of the support of a smoothed version of the real and synthetic data distributions. Their approximation is based on results of \citet[Method~IV]{confidence_sets_for_persistence_diagrams}, where it is argued that persistence diagrams from superlevel set persistence modules may carry topological information from the support of the data distribution. 
Specifically, the supports of the real and synthetic data are approximated by the superlevel sets $(\widehat{p}_{h_r})^{-1}[c_r, \infty)$ and $(\widehat{p}_{h_s})^{-1}[c_s, \infty)$, respectively, where $\widehat{p}_{h_r}$ and $\widehat{p}_{h_s}$ are KDEs for real and synthetic datasets $\mathcal{D}_r$ and $\mathcal{D}_s$ given by
\begin{equation*}
    \widehat{p}_{h_\bullet}(x)=\frac{1}{\left|\mathcal{D_\bullet}\right|}\sum_{p\in\mathcal{D}_\bullet}\frac{1}{h^d}\,K\left(\frac{x-p}{h}\right),
\end{equation*}
where $h>0$ and $K$ are the bandwidth and kernel of the KDEs,  selected beforehand, and  $c_r$ and $c_s$ are confidence bands for a given significance value $\alpha$, obtained using bootstrap over the datasets $\mathcal{D}_r$ and $\mathcal{D}_s$ with the condition that
\begin{equation*}
\liminf_{\left|\mathcal{D}_{\bullet}\right|\to \infty}\;\mathbb P\left(\lVert \widehat{p}_{h_\bullet}-p_{h_\bullet}\rVert_\infty < c_\bullet\right) \geq 1-\alpha,
\end{equation*}
where $p_{h_r}$ and $p_{h_s}$ are smoothed versions of the real and synthetic data distributions, respectively~\citep[Equation~(26)]{confidence_sets_for_persistence_diagrams}. By the stability theorems~\citep{stability_persistence}, the confidence bands $c_r$ and $c_s$, also bound the distances between the persistence diagrams coming from the functions $\widehat{p}_{h_\bullet}$ and $p_{h_\bullet}$ with probability $1-\alpha$, allowing one to study which points in the persistence diagrams generated from the data samples $\mathcal{D}_\bullet$ by $\widehat{p}_{h_\bullet}$ can be considered noise with respect to the persistence diagrams generated by $p_{h_\bullet}$; see~\citet[Section~4]{confidence_sets_for_persistence_diagrams}. The superlevel sets $(\widehat{p}_{h_\bullet})^{-1}[c_\bullet, \infty)$ are regions which intend to induce persistence diagrams without noisy points according to the previous confidence bands, thus recovering supports with similar topology to the topology of the support of the smoothed data distributions measured by the distances between their respective persistence diagrams. 

Under some technical assumptions, the precision and recall scores approximated using the superlevel sets proposed by \citeauthor{kim2023toppr}\ 
become close to the real precision and recall from the distribution as more examples are added to the datasets $\mathcal{D}_r$ and $\mathcal{D}_s$. The approximation is done with robustness, meaning that it holds even with data possibly corrupted by noise. The suitability of the new metrics was tested on several synthetic and real scenarios, where the metric accurately describes the differences between real and synthetic data distributions in several scenarios where other metrics struggle to evaluate  differences. Also, the F1-score computed from the precision and recall approximations ranked different generative models such as StyleGAN2~\citep{Karras2019stylegan2}, ReACGAN~\citep{kang2021rebooting}, among others, as the FID metric, the primary metric to score generative models in the moment of the publication of this survey, making this metric consistent with the state-of-the-art knowledge about the quality of generative models.

A class of generative models create synthetic data transforming values from a \textit{latent space} $\mathcal{Z}$ to the real data space $\mathcal{X}$ via a function $G\colon\mathcal{Z}\to\mathcal{X}$. 
For such generative models, a desirable property is to be able to control the attributes of the generated samples. One approach to controlling attributes involves designing generative models that map from the latent space to the data space in such a way that the latent space $\mathcal{Z}$ can be factored into subspaces corresponding to \textit{factors of variation} of the data generated, meaning that if one changes the value of a subspace associated with a factor of variation in the latent space, then the generated data change only in this source of variation. A model that satisfies this property is said to be \textit{disentangled}. For a formal discussion of disentanglement, we refer the reader to~\cite{higgins2018definition}. 
Preliminary work on disentangling generative models using topological data analysis to build regularization terms can be found in~\cite{balabin2023disentanglement}. In this work, regularization terms are based on the differences between generated data from two samples from the latent space using the representation topology divergence, a method to compare point clouds topologically
which is reviewed in Section~\ref{scn:pers_hom_internal_representations}.

Measuring a model's entanglement is challenging, and there is no canonical way to do~it. \cite{zhou2021evaluating} propose two measures of disentanglement, one unsupervised and one supervised, using topological data analysis. Although they perform similarly to other disentanglement metrics, these measures do not require some strong assumptions about the dataset or the model, except for the (weak) hypothesis that $\text{Im}(G)$ is a manifold for the unsupervised case and the more unrealistic assumption that the data space is a manifold for the supervised case. Roughly, these measures are based on two fundamental ideas for disentangled models:
\begin{enumerate*}
    \item The family of submanifolds of $\text{Im}(G)$ yielded by the images of $G$ restricted to the different values of the same subspace corresponding to a factor of variation are pairwise homeomorphic and thus have the same topology;
    \item Submanifolds corresponding to different factors of variation are usually not homeomorphic. 
\end{enumerate*}

The measures, based on the above ideas, penalize inconsistent topologies from the same factor of variation but reward topological differences between separate factors. However, the association of the latent dimensions of $\mathcal Z$ with the factors of variation is often unknown. Also, in practice, it is impossible to compute the full submanifolds for each restriction. To address this, the topological differences between the submanifolds of the different factors of variation are approximated by the topological differences of the submanifolds induced by the latent variables. Specifically, submanifolds are computed fixing values for the latent variables instead of for the factors of variation, and the topology for each submanifold is characterized by relative living times, that only need samples from the submanifolds. Then, Wasserstein barycenters~\citep{wasserstein_barycenters} are calculated for each latent dimension from the relative leaving times, leading to a dissimilarity matrix $M$ based on the pairwise Wasserstein distances between the Wasserstein barycenters. This matrix, processed by a (co)clustering algorithm, produces another dissimilarity matrix $M'$ measuring the \textit{topological similarities} of $c$ different probable factors of variation, from which the unsupervised disentanglement measure is derived as
\begin{equation}\label{eq:disentanglement_formula}
    \mu = \text{tr}(M')-\Bigg(\sum_{i=1}^c\sum_{j=1}^c M'_{i,j}-\text{tr}(M')\Bigg).
\end{equation} 
The supervised version, assuming that the data space is already factored into subspaces corresponding to the factors of variation, differs by generating the matrix $M$, this time not necessarily squared, as a Wasserstein dissimilarity matrix between the Wasserstein barycenters of the latent variables and of the factored real data. Finally, the supervised metric is calculated as in Equation~\eqref{eq:disentanglement_formula} for the leading principal submatrix of $M'$, generated from $M$ as before, taking the first $c$ rows and columns.

This metric was compared to other state-of-the-art disentanglement measures on three different datasets, dSprites~\citep{dsprites17}, a dataset to specifically test disentanglement of generative models, CelebA, and Celeba-HQ~\citep{karras2018progressive}, for ten different architectures including the aforementioned VAE and WGAN-GP, among others. In particular, it was shown that both, the unsupervised and the supervised proposed disentanglement metrics ranked models similarly to the other reference disentanglement metrics.

\subsection{Internal representations and activations}\label{scn:internal_representations_and_activations}
Most of the research presented in this survey belongs to this section. Here, we mainly focus on neural network weights and activations. Weights $\theta(\mathcal{N})$ are arguably one of the most important parts of neural networks, as they determine their functions alongside architectures $a(\mathcal{N})$. Therefore, selecting the appropriate weights when training is key to ensuring that neural networks perform effectively. On the other hand, activations are the result of the computations performed by the neural network $\mathcal{N}$ on the input data $x\in\mathcal{X}$, and therefore are key to understand the behavior of neural networks and their properties.

The use of TDA to analyze weights and activations is mainly divided into two different approaches. The first one, more qualitatively, is the use of Mapper, sometimes supported by some (persistent) homological computations, to understand the structure of the internal representations of neural networks. The second, more quantitatively, is the use of (persistent) homology to extract topological features of the internal representations of neural networks and link them with different properties of the networks, such as their generalization capabilities.

\subsubsection{Mapper}\label{scn:mapper_gtda}

This subsection is organized into two categories: first, those articles that apply Mapper to point clouds derived from weights, and second, those that utilize Mapper on point clouds resulting from activations.

\category{Mapper on weights}

\begin{figure}[t]
\centering
\includegraphics[width=0.75\textwidth]{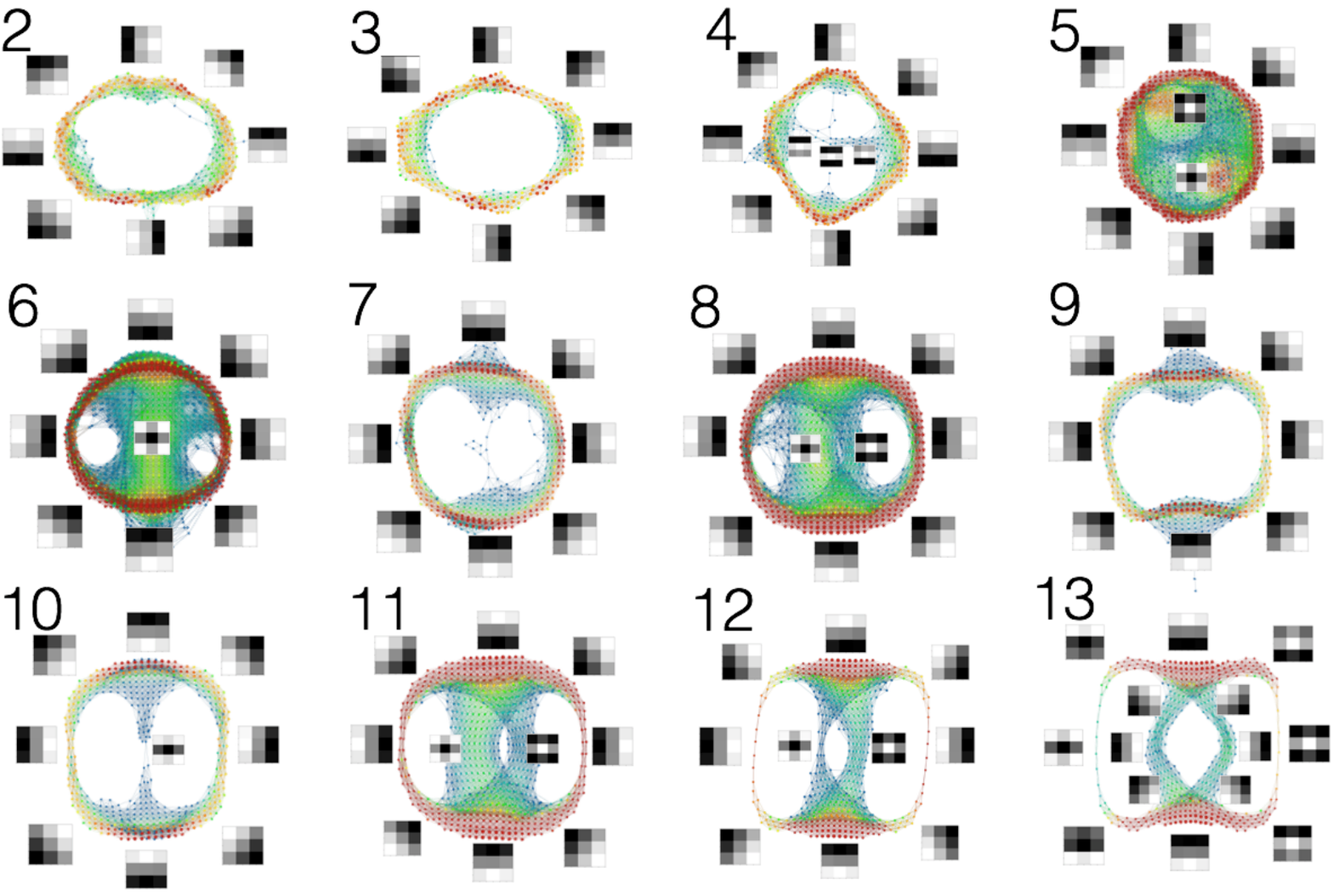}
\vspace*{-0.5cm}
\caption{Mapper graphs reproduced with permission from~\cite{exposition_and_interpretation_of_the_topology_of_neural_networks} of the convolutional filter weights from layer~2 to layer~13
of a trained VGG-16 neural network. Vertex colors represent the size of the collection represented by each node, increasing from blue to red. Matrices surrounding the Mapper graph are averages of filter weights located at nodes near the matrix, representing filters of 
the corresponding area of the graph.}
\label{fig:VGG-16Mapper}
\end{figure}
\noindent Two seminal papers by \citeauthor{topological_approaches_to_deep_learning}\, published in \citeyear{exposition_and_interpretation_of_the_topology_of_neural_networks} and a follow-up in \citeyear{topological_approaches_to_deep_learning}, stand as early explorations of the use of Mapper to analyze internal representations of FCFNNs, specifically CNNs. These papers focus on the topology and distribution of convolutional filter weights within the same layer of CNNs.

Consider a CNN with architecture $a$. Let $i$ be a convolutional layer containing filters $C^{(i)}_1,\hdots, C^{(i)}_{c^{(i)}}$ each of dimensions $h\times w\times c^{(i-1)}$. Here, $h\times w$ denotes the size of the convolution, $c^{(i-1)}$ denotes the number of channels in layer $i-1$, and $c^{(i)}$ denotes the number of channels in layer $i$ after convolution. Given $t$ independently trained instances of this architecture, Mapper graphs are generated for a subset of the $t\cdot c^{(i-1)}\cdot c^{(i)}$ vectors of dimension $h\times w$ derived from the convolutions across all $c^{(i-1)}$ channels and $c^{(i)}$ filters. 

In simple CNN architectures, comprising two convolutions, two pooling layers, and a fully connected output layer trained on the MNIST dataset, the Mapper graphs of the first convolutional layer closely resembled topological models of $3\times 3$ patches of natural images that were previously identified by~\cite{topology_of_natural_images}. This suggests that CNNs may capture inherent topological features of data during learning. Similar experiments were carried out on the CIFAR-10 dataset using architectures similar to those used for MNIST. Here, the results varied: some Mapper graphs resembled previously discovered topological models, while others exhibited new Mapper configurations. Mapper graphs of more advanced architectures VGG-16 and VGG-19, pretrained on ImageNet, were also analyzed. For example, in VGG-16, all convolutional layers except the first and thirteenth ones displayed a circle as the dominant topological structure, consistent with findings in natural image patches. Also, it was observed that in the first layers of the network, the topological structures were simpler than in the other ones. Figure~\ref{fig:VGG-16Mapper} displays the Mapper graphs for all convolutional layers in VGG-16, except the first one.

These Mapper graphs were related to generalization. On the one hand, more powerful networks appeared to learn simpler topological structures in the first layers, as seen in VGG-16, suggesting a possible correlation between the topological complexity of the network's weights and its generalization capabilities. On the other hand, constraints on the parameter space of neural networks based on previously identified weight topological structures improved network generalization. These topological structures also helped preprocess the input data to obtain an increase in accuracy without modifying the network. 

Building on the previous insights and the success of convolutional neural networks, \citeauthor{topological_approaches_to_deep_learning}\ proposed three ways to generalize convolutions to layers that take into account the topology of the data. This is continued in~\cite{love2020topological}, where topological CNNs, which encompass several topologically defined convolutional methods, are introduced. 

\cite{topology_of_learning_in_feedforward_neural_networks} proposed to use Mapper to study the evolution of weights through the training process for a fixed layer of a FCFNN. In this case, Mapper graphs were built using the \texttt{DBSCAN} clustering algorithm. In addition, two different filter functions were used: the $l^2$ norm and the projection to the three principal components of the Mapper input points. Given a FCFNN trained in $n$ steps and a layer $l$ with $N_l$ neurons, Mapper simplicial complexes were computed with a point cloud of $n\cdot N_l$ points of dimension $N_{l-1}$, each point representing the weights of the $N_{l-1}$ incident edges to a neuron of the layer $l$ at a given step of the training process. 

In a first experiment, \citeauthor{topology_of_learning_in_feedforward_neural_networks}\ trained an FCFNN in MNIST with only one hidden layer of 100 neurons, setting the bias values to zero and initializing the weights to zero. Using the $l^2$ function, the Mapper graph produced with the $l^2$ function and the output layer had ten different branches, the same number as the number of classes in the dataset. For the hidden layer, only twelve distinct branches were produced, much less than the number of neurons. This suggests that the number of branches in this Mapper graph configuration could capture the \textit{real expressiveness} of a neural network layer ---in this context, the real expressiveness of a neural network layer is a loose term. With it, we refer to a \textit{measure} quantifying the differences between the functions induced by each neuron on the layer~$i$. The higher the differences in activations for the different neurons in the layer, the higher this \textit{real expressiveness}. Furthermore, the branching patterns in the Mapper graph were correlated with the accuracy of the model in each training phase, supporting the previous claim. 

In a second experiment, performed with an FCFNN with two hidden layers, each containing 100 neurons, and weights initialized randomly following a normal distribution, the Mapper graph for the output layer also contained ten branches. However, for the first hidden layer, PCA projections of the weights studied seemed to distribute in a tree evolving in parallel on top of a smooth surface, and a Mapper surface generated using the first three principal components of the points further confirmed this fact. This surface was remarkably robust under variations in the training parameters and the initial weights. 

\category{Mapper on activations}

\noindent Mapper has also been used to analyze the structure of neural network activations. TopoAct, by~\cite{topoact}, was proposed as a visual exploration system to study the topology of activation vectors in fixed layers of neural networks. TopoAct uses the $l^2$ norms of the activation vectors and \texttt{DBSCAN} as the filter function and the clustering algorithm, respectively. The image cover of the filter function is built using an algorithm considering a fixed number of cover sets with a fixed percentage of overlap between them. Finally, each Mapper node is augmented with some feature visualization techniques or statistical data to obtain more insight into the activations that comprise it. 

One of the visualization techniques proposed by~\citeauthor{topoact}\ was the computation of \textit{activation images} for the activations collected in each node and for its average. Activation images, given an activation vector or an average of them, are \textit{idealized} images that would have produced the activations via an iterative optimization process proposed by~\cite{olah2017feature, olah2018the}. Other statistical data proposed by~\cite{experimental_observations_of_the_topology_of_cnns} to describe Mapper nodes were:
\begin{enumerate*}
    \item Pie charts showing the composition of class labels in that node, associating to each activation vector the label of the image that generated the activation vector;
    \item Node-wise purity, defined as $\alpha_i= c_i^{-1}$ where $c_i$ is the number of class labels in the Mapper node~$i$;
    \item Class-wise purity, defined as $\gamma_y= \left|J_y\right|^{-1}\sum_{x\in J_y}\beta_x$, where $J_y$ is the set of activations in the Mapper graph associated to the class $y$ and $\beta_x=\left|I_x\right|^{-1}\sum_{i\in I_x}\alpha_i$, with $I_x$ being the set of nodes in the Mapper graph containing the activation $x$.
\end{enumerate*}

The construction method for activation vectors varies depending on the type of layer under consideration. In particular, the convolutional layer emerged as the most extensively studied using TopoAct among the articles analyzed in this survey. For convolutional layers with output values of dimension $h\times w\times c$, \cite{topoact} proposed to compute activations for each input value by randomly selecting two indices $i, j\in[h]\times[w]$ and taking the $c$ dimensional vector obtained by fixing the dimensions $i, j$ of the output. Later, two other sampling methods for activation vectors were proposed by~\cite{experimental_observations_of_the_topology_of_cnns}: 
\begin{enumerate*}
    \item Sampling all the possible $h\cdot w$ activations vectors for all possible pairs of indices $i, j$;
    \item Sampling activations such that the indices $i,j$ whose receptive fields are associated with the most quantity of background or foreground pixels in the inputs are the ones selected. 
\end{enumerate*}

For pretrained versions of InceptionV1~\citep{going_deeper_with_convolutions}, BERT~\citep{bert}, and ResNet-18~\citep{resnets} the Mapper graphs generated by TopoAct carried meaningful information. For InceptionV1, \citeauthor{topoact}\ observed that branches in the Mapper graphs corresponded to activations containing different features of the inputs. For example, the activation images for one bifurcation showed that the nodes of one branch were associated with animal legs, while the nodes of the other were associated with distorted faces. Similarly, loops of nodes were associated with different aspects and features of the same underlying objects. An example of this was found in a loop that contained six nodes, each representing different features (body parts) of some set of animals, including dogs and foxes. For BERT, a language representation model, the Mapper graphs showed a similar behavior separating word representations, where branches seemed to separate contextual meaning of similar words. 

For ResNet-18, \citeauthor{topoact}\ observed branching patterns similar to those seen in the InceptionV1 model, suggesting that TopoAct results are not specific to a particular dataset or architecture. Furthermore, \citeauthor{experimental_observations_of_the_topology_of_cnns}\ observed that the complete and random activation sampling methods yielded similar bifurcation patterns. For the sampling method associated with the most background or foreground pixels in the input images, notable class bifurcations, that is, branches in the graph separating activations associated with different labels, seemed to appear at earlier layers, although this is not evident from the experiments.  In general, node-wise and class-wise purity were found to be higher in deeper layers for all the sampling methods, confirming the idea that models get better at separating the classes the deeper they go. 

In a more quantitative way, \citeauthor{experimental_observations_of_the_topology_of_cnns}\ also proposed a dissimilarity measure to quantitatively compare two layers. This dissimilarity is given by the sliced Wasserstein distances of the persistence diagrams induced by samples of the activations of both layers. Specifically, the activations with the highest $l^2$ norms are taken. However, although the measure was robust to different weight initializations, the dissimilarity did not pass some \textit{ sensitivity tests} proposed by~\citep{ding2021grounding} that reasonable measures between layers should pass.

TopoAct has also been used by~\cite{zhou2023visualizing} to analyze the effect of adversarial examples in neural networks using the Mapper graphs of one of their deepest layers computed with the training dataset. Specifically, Mapper graphs are studied using a variant of the node-wise purity introduced in~\cite{topobert}.  For a node $i$, the purity of $i$ is $1-H(D_i)/H(D)$ where $H$ denotes the Shannon entropy of a distribution, $D_i$ denotes the observed distribution of labels for the points in the node $i$, and $D$ denotes a uniform distribution of all labels. This purity reaches one whenever all points in $X$ are of the same class and zero when points are distributed uniformly over all classes. \citeauthor{zhou2023visualizing}\ studied how performing adversarial trainings, i.e., training neural networks with adversarial examples, affects the configuration of the Mapper graph. The experiments were performed in two scenarios: 
\begin{enumerate*}
    \item Training a simple FCFNN model with MNIST;
    \item Training a ResNet-18 with CIFAR-10.
\end{enumerate*}

In the first case, where no overfitting was observed during training, impure nodes, that is, nodes that do not contain a dominant label in them, of the neural network trained without adversarial examples captured decision boundaries. Also, the higher the attack, that is, the bigger the perturbations made to the original examples, the higher the number of nodes with low accuracy in the Mapper graphs, i.e., the higher the number of nodes containing activations coming from inputs that were misclassified.

In the second case, where overfitting was observed, the higher the attack, the lower the weighted average purity, and the lower the test accuracy, where the purity is weighted by the number of activations in each node divided by the total number of activations in the Mapper graph. The observation that the model is overfitting and that the purity decreases with the attack suggests that, in this case, the impure nodes were also capturing decision boundaries.

Based on such observations, \citeauthor{zhou2023visualizing}\ propose to improve the robustness of adversarially trained neural networks by selecting misclassified activations from low-accuracy nodes on the mapper graph and using them to refine the model. For the FCFNN network in MNIST, this procedure marginally improved the accuracy of the model. However, for the ResNet-18 in CIFAR-10, this procedure had no effect.

\subsubsection{Homology and persistent homology}\label{scn:pers_hom_internal_representations}

The literature has a wealth of methods to leverage the information provided by (persistent) homology features induced by activations and weights of neural networks. We divide this section into six blocks, each representing a space of internal representations in which TDA is applied, as follows:
\begin{enumerate*}
    \item Activations in the complete neural network graph;
    \item Activations for each layer;
    \item Weights in the complete neural network graph;
    \item Weights layer by layer;
    \item Activations whose dissimilarities depend on the weights; and
    \item Generic spaces.
\end{enumerate*}

\category{Activations in the complete neural network graph}

\noindent The (persistent) homology of activation vectors has been successfully used to analyze many aspects of deep neural networks, such as their generalization or interpretability. In an early study within this section, \cite{whatdoesitmean_corneanu} 
analyzed how diverse topological information extracted from the set of neuron activations of neural networks was correlated with their generalization capabilities. The neuron activations were studied for the complete graph at the same time, and the activation vector $a_{v_i^l}$ for each neuron $v_i^l$ was taken as
\begin{equation*}
a_{v_i^l}=\left(\phi_\mathcal{N}^{(l)}(x_1)_i,\hdots, \phi_\mathcal{N}^{(l)}(x_n)_i\right),
\end{equation*}
for a fixed set of inputs $\mathcal{D}=\{x_i\}_{i=1}^n$ to the neural network $\mathcal{N}$, in this article a (sub)set of the training dataset.

In their first experiments, \citeauthor{whatdoesitmean_corneanu}\ studied how the number of simplices of Vietoris--Rips simplicial complexes, given by the formula
\begin{equation*}
    S(n)=\left|\sigma\in \text{VR}_t(P, d_{\downarrow}):\text{dim}(\sigma)=n\right|,
\end{equation*}
varied during several training processes of a LeNet neural network architecture~\citep{leNet}. Given a neural network $\mathcal{N}$ and a fixed value $T$, Vietoris--Rips simplicial complexes were computed from point clouds $(P, d_{\downarrow})$ induced from a weighted connected graph $\mathcal{F}_\mathcal{N}$ such that:
\begin{enumerate*}
    \item The vertices are a subset of neurons of $\mathcal{N}$;
    \item Edges are weighted by $w_E(\{v, w\})=\left|\text{corr}(a_{v}, a_{w})\right|$, where $\text{corr}$ is the sample Pearson correlation and $a_{v}$ and $a_{w}$ are the activation vectors of the neuron $v$ and $w$, respectively;
    \item An edge $\{v,w\}$ is in $E(\mathcal{F}_\mathcal{N})$ if $w_E(\{v, w\}) > T$;
    \item Only edge endpoints are added to the vertex set.
\end{enumerate*} 
The parameter $t$ for the Vietoris--Rips simplicial complexes was chosen such that the density $\rho_t$ of the edges with non-zero correlation for the parameter~$t$, given by
\begin{equation*}
    \rho_t = \frac{\left|\{\sigma=\{v, w\} \in \text{VR}_t(P, d_{\downarrow}): \text{dim}(\sigma)=1 \text{ and }\left|\text{corr}(a_v, a_w)\right|>0\}\right|}{\left\{\sigma=\{v, w\}\in\left|E(\mathcal{F}_\mathcal{N}): \left|\text{corr}(a_v, a_w)\right|>0\right|\right\}},
\end{equation*}
was near to $0.25$. In the experiments, it was found that, the higher the area under the curve of $S(n)$, the better the generalization capabilities of the neural network. In particular, it was shown that, for such training procedures, the function $S(n)$ is capable of distinguishing between the three main regimes given during training: underfitting, generalization, and overfitting. That is, the training process starts with a small area under $S(n)$ (underfitting regime), then $S(n)$ grows and achieves its maximum value (generalization regime), and then decreases again (overfitting regime).

The previous results motivated more involved experiments to see the relationship between the topological properties of the activations of neural networks and their generalization capacity. In particular, a new functional graph $\mathfrak{F}_\mathcal{N}$ is built to induce Vietoris--Rips filtrations to extract sharper topological information about the network. In this case, the functional graph is complete, its set of vertices is a fixed (sub)set of neurons of $\mathcal{N}$, and its weights are given by a correlation dissimilarity $w_E(\{v,w\})=1-\left|\text{corr}(a_v, a_w)\right|$.  Here $\mathfrak{F}_\mathcal{N}$ induces point clouds $(P,d)$ of neurons where $P=V(\mathfrak{F}_\mathcal{N})$ and $d=w_E$. From this point cloud $(P,d)$, generated again during several training procedures of a LeNet neural network, \citeauthor{whatdoesitmean_corneanu}\ computed persistence diagrams $D(\mathbb V_k(\text{VR}(P,d)))$ for $k\in\{1,2,3\}$ and a normalized version of their Betti curves $b_k(t)=\left|\left\{(b,d)\in D(\mathbb V_k(\text{VR}(P,d))):t\in [b, d)\right\}\right|$ to study the evolution of the topology during training again.

During the first epochs, it was observed that the highest values of these curves moved from the left part of the domain to the right for $k\in[3]$. However, when the training entered into the overfitting regime, the maximum values of the normalized Betti curves moved again to the left part. Due to the behaviour of the normalized Betti curves, \citeauthor{whatdoesitmean_corneanu}\ proposed an early stopping of the training whenever it started moving the maximum values of Betti curves to the left again. Also, \citeauthor{whatdoesitmean_corneanu}\ used normalized Betti curves to detect sets of adversarial examples, if any. 

The observed evolution of Betti curves suggests that the topological correlation structure of the activations of neural networks is highly related with their capacity to generalize to the whole data distribution. This was further studied by \cite{computing_test_error_corneanu} and \cite{ballester2023predicting}, 
who used persistence diagrams $D(\mathbb V_k(\text{VR}(P,d)))$ of dimensions $k\in\{0, 1\}$ to predict the generalization gap, that is, the difference between training and test accuracies. 

First, \citeauthor{computing_test_error_corneanu}\ proposed to predict the generalization gap by performing a linear regression with independent variables chosen to be two persistence summaries,
namely average persistence $\lambda$ and average midlife $\mu$, given by
\begin{equation*}
    \lambda(D) = \frac{1}{\left|D\right|}\sum_{(b,d)\in D}d-b,\qquad \mu(D) = \frac{1}{\left|D\right|}\sum_{(b,d)\in D}\frac{d+b}{2},
\end{equation*}
respectively. This approach seemed to work on controlled computer vision scenarios. Following these results, \cite{ballester2023predicting} extended previous work by studying more networks with different generalization gaps extracted from the first and second tasks of the \textit{Predicting Generalization in Deep Learning} NeurIPS challenge~\citep{jiang2020neurips}. In this study, more persistence summaries were tested, such as persistent entropy~\citep{persistent_entropy}, persistence pooling vectors~\citep{persistence_pooling}, and complex polynomials~\citep{complex_polynomials}, among others. 

The problem with the aforementioned approach is that it does not scale properly for the more complex networks analyzed in~\cite{ballester2023predicting}. Due to the high number of neurons in larger neural networks and the high number of training examples in the datasets, computing persistence diagrams for the whole set of neurons and training examples was unfeasible. To alleviate this problem, \citeauthor{ballester2023predicting}\ proposed to generate multiple persistence diagrams from uniform samples of the dataset and from neuron samples using a probability distribution where neurons with higher activations in absolute value had a higher chance of being sampled than neurons with lower activations. From the corresponding persistence diagrams, persistence summaries are computed and then bootstrapped to perform linear regression to predict the generalization gap. 

The best persistence summaries to predict generalization gaps in this line of work were a combination of statistical measures ---such as averages and standard deviations--- of the points in persistence diagrams. In particular, fixing some sources of variability of the networks affecting their depth, \citeauthor{ballester2023predicting}\ found a correlation between the averages and standard deviations of the second coordinates of zero- and one-dimensional persistence diagram points and the generalization gap of their associated networks. These insights and methods were further studied and used in a follow-up article by~\cite{ballester2023decorrelating} to develop regularization terms that minimize pairwise correlations between neurons. 

A very similar approach was used to detect trojaned neural networks. A brief explanation of trojaned neural networks can be found in Section~\ref{scn:trojaned_neural_networks}. \cite{topo_detection_trojaned} propose to build zero- and one-dimensional Vietoris--Rips persistence modules from activations as in the previous works but replacing the Pearson correlation coefficient in absolute value by a more general correlation coefficient measure. For such persistence modules, \citeauthor{topo_detection_trojaned}\ found a significant difference between clean and trojaned neural network persistence diagrams for both synthetic and real-world experiments, where the real-world experiments comprised training 70 ResNet18 using a clean MNIST dataset and another 70 with a trojaned MNIST counterpart. The average death time of zero persistence diagrams was particularly significant for segregating persistence diagrams from clean and trojaned neural networks, where averages were significantly lower than those of clean models. Also, by manual inspection of cycle representatives of the points of persistence diagrams, it was observed that trojaned models contained edges connecting shallow and deep layers, a phenomenon that did not happen for clean models. With this in mind, \citeauthor{topo_detection_trojaned}\ trained basic FCFNN models to detect trojaned convolutional neural networks trained on MNIST, CIFAR10 and the IARPA/NIST TrojAI competition~\citep{trojai_dataset} datasets using persistence summaries from the persistence diagrams of the networks as input. These models resulted in higher performance than those for other state-of-the-art trojan detection algorithms and models.

\category{Activations for each layer}

\noindent While in previous works the topology of the activations was studied globally, \cite{topology_of_deep_neural_networks} studied the evolution of the activations layer by layer in FCFNNs for binary classification problems in an extensive set of experiments. For data samples $\mathcal{D}_i$ from one of the classes $i\in[2]$ at a time, the evolution of the (persistent) homology of $\mathcal{D}_i$ through the different layers is analyzed. In particular, for each example $x\in\mathcal{D}_i$ and each layer $l$ in an FCFNN $\mathcal{N}$, an activation vector $a_x^l$ is computed by taking the vector composed of the different activation values of the neurons given the example $x$ in layer~$l$, that is, 
\begin{equation*}
   a_x^l=\left(\phi_\mathcal N^{(l)}(x)_1, \hdots, \phi_\mathcal N^{(l)}(x)_{N_l}\right). 
\end{equation*}
Then, persistent homology is computed from Vietoris--Rips filtrations of the point clouds $\mathcal{D}_i^l=\left\{a_x^l:x\in\mathcal{D}_i\right\}$ for each layer~$l$. 

The experiments were divided into two different scenarios: synthetic and real datasets. For the first scenario, the datasets were samples from spaces with known topology. For the second scenario, the datasets were simplified and binarized classification versions of several known datasets, such as MNIST. In both cases, FCFNNs were trained with different architectures, different training parameters, and almost zero training error.

For the synthetic datasets, the evolution of Betti numbers was studied for different dimensions $k$ of Vietoris--Rips complexes of fixed value~$t$. The dissimilarity for these complexes was chosen to be the graph geodesic distance on the $k$-nearest neighbors graph applied to the point clouds $\mathcal{D}_i^l$. For a fixed sample $\mathcal{D}_i$ and a fixed neural network $\mathcal{N}$, the parameters $t$ and $k$ were selected in such a way that $b_0(\text{VR}_t(\mathcal{D}_i))$, $b_1(\text{VR}_t(\mathcal{D}_i))$, and $b_2(\text{VR}_t(\mathcal{D}_i))$ were equal to the first three (known) Betti numbers of the space from which the dataset $\mathcal{D}_i$ was sampled. However, due to the difficulty in selecting good parameters $t$ and $k$ for the real data, persistence diagrams of the activations for each layer were instead studied in the second scenario.

For the synthetic scenario, \citeauthor{topology_of_deep_neural_networks}\ observed a decay of the different Betti numbers $b_k$ through the layers. This decay was slower through different experiments for completely smooth functions than for ReLU-like (ReLU and leaky ReLU) activation functions for dimension zero. The number of layers needed to perform this topological simplification varied according to the topology of the initial data. Narrow layers, that is, those with few neurons per layer, appeared to simplify the topology faster than their wider counterparts. Bottleneck architectures, i.e., architectures with decreasing numbers of neurons per layer, seemed to force larger topological changes in the data than networks with constant number of neurons per layer. Also, depth did not seem to influence the way the topological simplifications are distributed across the layers: initial layers did not seem to perform many topological simplifications, in general. Thus, reducing depth simply concentrated the topological simplifications in the last layers. This topological simplification was also observed in the persistent diagrams of the activations for the real datasets in simpler networks than the ones used for the synthetic scenario, where the number of points and their persistence across all dimensions diminished through layers.

The previous observations were contradicted by~\cite{activation_landscapes_as_a_topo_summary}. In this work, evolution of the topology of the activations through the different layers of FCFNNs for synthetic and real data was studied again. However, this time, the activations and distances between them were preprocessed and slightly modified depending on the experiments performed to build a point cloud. From such preprocessed point clouds, Vietoris--Rips persistence diagrams were computed using the usual Euclidean distance between vectors. Finally, instead of studying Betti numbers or persistence diagrams directly, \citeauthor{activation_landscapes_as_a_topo_summary}\ computed persistence landscapes $\lambda^l_k=\big(\lambda^l_{k, 1}, \lambda^l_{k, 2}, \hdots\big)$ derived from the persistence diagrams for each layer for different homological degrees~$k$.

In the previous work, Betti numbers were used to quantify the \textit{topological complexity} of activations in a given layer. Persistence landscapes also allow one to define topological complexities over point clouds. 
Specifically, given a persistence landscape $\lambda$ coming from the point cloud, one way to measure its complexity is to measure the area under its curves~$\lambda_i$. The sum of these areas defines an inner product
\begin{equation*}
    \left\langle\lambda, \lambda'\right\rangle =\sum_{i=1}^\infty\int_{-\infty}^\infty\lambda_i(t)\lambda'_i(t)\,dt,
\end{equation*}
that induces a norm $\left\|\lambda\right\|=\sqrt{\left\langle\lambda, \lambda\right\rangle}$. The higher the norm, the higher the topological complexity of the point cloud. This topological complexity measure was the one used by \citeauthor{activation_landscapes_as_a_topo_summary}\ to measure the topological evolution of the activations computing it layer by layer, fixing a homological degree~$k$.

The results for synthetic data were similar to the previous ones: for FCFNNs with $11$ layers with ReLU activation functions except for the output one, trained to perfect accuracy with $100$ different initialization weights, topological complexities were observed to decay through the last layers, where the best weight configurations decayed faster on average than their worse counterparts. However, for the first layers, the topological complexities increased layer by layer. More strikingly, for FCFNNs trained to near-perfect accuracy with $7$ layers distributed following a bottleneck architecture with ReLU activation functions except for the output one, topological complexities were observed to increase in the last layers on average and only decreased in the first layers. Observations on these two different architectures question the previous results by~\citeauthor{topology_of_deep_neural_networks}\ and motivate further research on the topic. 

Another promising approach, deeply related to the study of the decision regions, is the analysis of the topology of the activations of the output layer. The fundamental hypothesis is similar to what was found in~\cite{topology_of_deep_neural_networks}: the easier the topology of the last layer, the more robust, and thus the better, the neural network. In this regard, \cite{experimental_stability_analysis} studied the zero and one-dimensional Vietoris--Rips persistence diagrams of the last-layer activations, taken as in the previous articles, using the usual Euclidean distance. Although they did not prove the main hypothesis and could not relate network performance with output layer topology, it is a first step towards this direction.

Intuitively, the topology of the activations for inputs of the same class must be similar. Based on this hypothesis, \cite{zhao2022quantitative} propose a way to measure the \textit{quality} of convolutional filters for individual channels of convolutional layers in neural networks for classification problems. In this work, the inputs are squared images, and we assume that the widths and heights of convolutional layer outputs are also equal. To compare the topology of the activations produced by the convolutional filters of interest, undirected weighted fully connected graphs are built. For a layer $l$, channel $c\in[c^{(l-1)}]$, and input sample $x$, the filter graph $C_x^{l, c}$ is generated such that $V(C_x^{l, c})= [h]$ and 
\begin{equation*}
   w_{E}(\{i, j\})= \max\left(\phi^{(l)}_\mathcal{N}(x)_{i, j, c}, \,\phi^{(i)}_\mathcal{N}(x)_{j, i, c}\right),
\end{equation*}
for $i, j\in[c^{(l-1)}]$, where $\phi^{(l)}_\mathcal{N}(x)_{j, i, c}$ is the $(i,j,c)$ output value of $\mathcal{N}$ for the $l$-th convolutional layer given~$x$. For each of these graphs, \citeauthor{zhao2022quantitative}\ computed the infimum of the support of the associated Betti curves coming from the persistence module $\mathbb V_1(\text{VR}^1(V(C_x^{l, c}), d_\downarrow^0))$, which is directly related to the time in which the first \textit{non-trivial cycle}, i.e., not given by a combination of triangles, appears in the filtration. We denote this minimum by $b_1^\text{inf}(x, l, c)$. The values $b_1^\text{inf}(x, l, c)$ induce discrete random variables $B_1(l, c, y)$ for each label $y$ of the classification problem with sample space $\Omega_y$ the subset of samples from the training dataset $\mathcal{D}_\text{train}$ with common label~$y$. These random variables have probability distributions 
\begin{equation*}
    P_{1, l, c, y}(n) = \frac{b_n}{\left|\Omega_y\right|}, \qquad b_n = \sum_{x\in\Omega_y}\mathbbm{1}_{\left\{x'\,:\; b_1^{\text{inf}}(x', l, c)=n\right\}}(x),
\end{equation*}
that capture information about the similarity of the topology of the activations given by the convolutional filters of interest in the data distributions of the different labels of the classification problem. This similarity can be further quantified using the entropies of the probability distributions, given by
\begin{equation}\label{eqn:feature_entropies}
    H'_{1, l, c, y} = -\sum_{n=0}^{\infty} P_{1, l, c, y}(n)\log P_{1, l, c, y}(n).
\end{equation}

The values given by Equation~\eqref{eqn:feature_entropies} are called \emph{feature entropies}, and are the topological summaries proposed by \citeauthor{zhao2022quantitative}\ to measure the quality of convolutional filters. Note that the infimum value $b_1^{\text{inf}}$ may not exist in some cases. For layers and channels in which this happens many times, the entropy approaches zero, although entropy is not really informative as it is affected by the non-existence of infimum values. In such cases, we modify entropy as
\begin{equation*}
    H_{1, l, c, y} = \begin{cases}
        H'_{1, l, c, y}&\text{if } \varepsilon_{1, l, c, y} \geq p,\\[0.1cm]
        (1-\varepsilon_{1, l, c, y})\log\left|\Omega_y\right|&\text{otherwise},
    \end{cases}
\end{equation*}
where $\varepsilon_{1, l, c, y}$ is the percentage of images in class $y$ having birth times and $p$ the minimum percentage of images we admit to use the real entropy, in the article, $p=0.1$. 

\citeauthor{zhao2022quantitative}\ demonstrated the effectivity of the entropy measure to obtain information on the \textit{quality} of the convolutions and of the entire neural network for VGG-16 models trained on the ImageNet dataset. They observed that, for well-trained neural networks, the feature entropy continually decreased as the layers went deeper, while for random weights this decreasement is absent. On the other hand, they observed that, during training, the feature entropies of the last convolutional layer decreased, and were highly correlated with the evolution of the cross entropy training loss, suggesting that feature entropies are good indicators of the generalization of networks. 

Feature entropies also were invariant to weights reescaling, a desirable property to measure the quality of the convolution operations, as reescalings of weights have no substantial impact on the network performance in general. Additionally, randomness of the weights was also detected by comparing the feature entropies of trained and randomly initialized networks. Finally, for the last convolutional layers, models with better generalization were connected to low feature entropies.

\category{Weights in the complete neural network graph}

\noindent Recall that the input values influence the output of a neural network via the different input-output paths available in the neural network graph. The influence of each path is characterized by the magnitudes of the weights associated with the edges in the path, which modify the magnitudes of the activations, and thus their relevance, at each step. \cite{topological_measurement_of_deep_networks} propose to build neural network graph filtrations taking into account the influence of the different paths in the network. 

Let $\mathcal{N}$ be a neural network, and $v_l^i, v_{l+1}^j\in V(G(\mathcal{N}))$ be two connected neurons in the directed graph $G(\mathcal{N})$. Formally, the relevance of the directed edge $(v_l^i, v_{l+1}^j)$ connecting $v_l^i$ and $v_{l+1}^j$ in the output calculation is defined to be the linearly rectified weight of the edge normalized by the sum of the linearly rectified weights of edges pointing to~$v_{l+1}^j$, that is, 
\begin{equation*}
    R_{v, w}=\frac{\text{ReLU}\big(W_{i,j}^{(l+1)}\big)}{\textstyle\sum_{k\neq i}\,\text{ReLU}\big(W_{k,j}^{(l+1)}\big)}.
\end{equation*}
Then, the influence from a vertex $v$ to a vertex $w$ is characterized as the maximum product of the relevances of the edges among all the paths between $v$ and $w$ whenever $v\neq w$, and $1$ otherwise, i.e.,
\begin{equation*}
    \tilde{R}_{v,w}=\begin{cases}
        \max_{(v, p_1, \hdots, p_n, w)\in P_{v,w}}R_{v, p_0}\left(\prod_{i=1}^{n-1}R_{p_i, p_{i+1}}\right)R_{p_n, w}&\text{if $ v\neq w$,}\\
        1 &\text{otherwise,} 
    \end{cases}
\end{equation*}
where $P_{v, w}$ is the set of all paths starting at $v$ and ending at $w$ in $G(\mathcal{N})$. 

The influences between vertices of a neural network induce a dissimilarity function over the neurons of the network, which can be used to build filtrations. However, to preserve information about edge directions, \citeauthor{topological_measurement_of_deep_networks}\ propose an alternative way to build filtrations $\left(K_i\right)_{i=1}^n$ from a monotonically decreasing sequence of indices $\left(t_i\right)_{i=1}^n$ given~by
\begin{equation*}
    K^p_i = K^p_{t_i} = \begin{cases}
        V(G(\mathcal{N}))&\text{ if }p=0,\\
        \{\{v_{k_0}.\hdots, v_{k_p}\}: \tilde{R}_{v_{k_i}, v_{k_j}}\geq t_i\text{ for all } k_i > k_j\}&\text{ if }p\geq 1,
    \end{cases}
\end{equation*}
where $p$ indicates the dimensions of the simplices of~$K^p_i$, and the vertices $V(G(\mathcal{N}))$ are ordered in such a way that if the layer number of $v_i$ is lower or equal than the layer number of $v_j$, then $i \geq j$.

The filtration $\left(K_i\right)_{i=1}^n$ captures the influence of the different paths between neurons in the neural network graph and can be easily extended to a persistence module where the simplicial complex assigned to the time $t\in\mathbb R$ is $K_{\lfloor t \rfloor}$ for $0\leq t \leq n$, $K_0$ for $t < 0$, and $K_n$ for $t > n$. Preliminary results about simple networks trained on MNIST and CIFAR-10 suggest that the distribution of points of one-dimensional persistence diagrams computed from persistence modules coming from the previous filtration are correlated with several properties of the network and the dataset, such as problem difficulty or network expressivity. For example, the appearance of points near the diagonal of the persistence diagrams was associated with a shortage of data of specific labels in the training dataset. In addition, persistence diagrams seemed to be robust with respect to the initial weight values of the neural network weights, yielding each architecture similar persistence diagrams for different initializations after training. However, a more thorough study of such persistence modules and their associated persistence diagrams is needed to understand their relationship with the generalization capabilities of a neural network, as \citeauthor{topological_measurement_of_deep_networks}\ remark.

An extension of the relevance quantity for convolution and pooling operations leads to persistence modules better suited to analyze convolutional neural networks. This extension was presented in~\cite{overfitting_measurement_CNNs_using_trained_network_weights} to study the overfitting of convolutional neural networks. \citeauthor{overfitting_measurement_CNNs_using_trained_network_weights}\ observed that, in simple scenarios, the distribution of the points in the one-dimensional persistence diagram was correlated with the overfitting of the network, as measured by the difference in train and test accuracies. Specifically, for each given architecture, the number of points near the diagonal increased according to the increase in the dropout rate of the network used in their experiments, which was correlated with the overfitting of the network. In addition, a high total number of points in the persistence diagram was correlated with less underfitting in neural networks. 

The observed correlation between points of persistence diagrams and generalization of neural networks make persistence diagrams potential tools for comparing neural networks according to their generalization capabilities. However, straightforward persistence diagrams were found not to be entirely suitable for comparing models with different architectures. To address this difficulty, \citeauthor{overfitting_measurement_CNNs_using_trained_network_weights} proposed a novel approach: pruning the networks of interest using a magnitude-based strategy \citep{neural_network_pruning} before generating persistence diagrams to compare them. After selecting several groups of neural networks to be compared, an overall high correlation between the number of points near the diagonal of normalized persistence diagrams and overfitting of the networks of the different groups was observed, validating the pruning method as a suitable way to generate normalized persistence diagrams to compare different models.

A slightly modified relevance function was used in filtrations by~\cite{network_pruning_using_PH} to perform edge pruning in neural networks. Using $R_{v, w}= \big|W_{i,j}^{(l+1)}\big|/\sum_{k\neq i}\big|W_{k,j}^{(l+1)}\big|$, \citeauthor{network_pruning_using_PH}\ proposed the following pruning algorithm: 
\begin{enumerate*}
    \item Sort points $(b,d)$ of one dimensional persistence diagrams by their value $b+d$ in ascending order;
    \item For each point, choose a cycle representative $c$ of the point $(b,d)$;
    \item Select the edges contained in the representatives in the previous step until you reach a desired number of edges to conserve;
    \item Prune the edges not selected in the previous step.
\end{enumerate*}
This method was shown to be competitive with respect to the global magnitude algorithm~\citep{neural_network_pruning} in terms of final accuracy of the pruned networks. 

A similar approach to the one taken by~\cite{computing_test_error_corneanu} and by~\cite{ballester2023predicting} was proposed by~\cite{on_the_use_of_ph_to_control_the_generalization_capacity_of_a_nn}. \citeauthor{on_the_use_of_ph_to_control_the_generalization_capacity_of_a_nn}\ argued that the dataset has too much influence on the activations and thus their correlations do not capture all the relevant information about the neural network related with generalization. Instead, they proposed to analyze zeroth persistence diagrams $D(V_0(\text{VR}(P, d)))$ induced by a subset $P$ of the vertices of a neural network $\mathcal{N}$ and dissimilarities given by the Euclidean distance between scores based on weights associated to each vertex. The \emph{score} associated to a node $v\in V(G(\mathcal{N}))$ in the network graph is the relevance value $S_v$ given by
\begin{equation*}
    S_v= \!\!\!\!\!\!\!\!\sum_{(v, w)\in E(G(\mathcal{N}))}\!\!\!\!\!\!\!\!\!\pi_{w, v}\cdot \delta_w, \quad \pi_{w,v}= \frac{\left|W_{w, v}\right|}{\sum_{(u, w)\in E(G(\mathcal{N}))} \left|W_{w, u}\right|},\quad \delta_w=\begin{cases}
        1 &\text{ if }w\in V_L,\\
        S_w &\text{ otherwise, }
    \end{cases}
\end{equation*}
where $W_{w,v}$ is the weight corresponding to the edge $(v,w)\in E(G(\mathcal{N}))$ and $V_L$ is the set of neurons of the output layer. This score is the HVS score for a neuron~\citep{yacoub1997hvs} which gives, for each neuron, a score of the neuron based on the magnitudes of the weights of its outcoming edges, that is related to the contribution of the neuron to the output value. The set $P$ of points used to computed persistence diagrams is simply the 90\% of neurons with highest relevance score. 

To study the relationship of these diagrams with the generalization gap, \citeauthor{on_the_use_of_ph_to_control_the_generalization_capacity_of_a_nn}\ trained very simple FCFNNs with two hidden layers for a variety of datasets from the UCI Machine Learning repository~\citep{uci_machine_learning_repo}. For the experiments, the generalization gap was compared with the average persistence value of the diagram generated at each iteration of the training procedure. Linear regression models were fitted with the generalization gaps and average persistences during training as dependent and independent variables, respectively. The $R^2$ values of these linear regressions were slightly greater than the ones for the same linear regression models using the persistence diagrams computed in~\cite{ballester2023decorrelating}. However, the simplicity in the experiments performed, both in the experimental pipeline and in the networks used, makes that further experimentation is needed to evaluate if this approach generalizes to more complex scenarios and also improves the methods of \citeauthor{ballester2023decorrelating}\ in them.

\category{Weights layer by layer}

\noindent Layerwise topological complexities have also been studied as a function of the weights. \cite{rieck2018neural} proposed to study the so-called neural persistence. Given a non-output layer $l$ of a neural network $\mathcal{N}$, its \emph{neural persistence} is defined as
\begin{equation*}
    \text{NP}_l(\mathcal{N})= \Bigg(\sum_{(b,d)\in D}\left|d-b\right|^p\Bigg)^{1/p},
\end{equation*}
where $D = D^w\left(\mathbb V_0\left(\text{VR}^1(V(G_l), d_{\downarrow}^V)\right)\right)$ is the zeroth persistence diagram of the weighted complete bipartite graph $G_l$ with vertices $V_l\sqcup V_{l+1}$ and weights given by $w_V(v) = w_\text{max}$ and $w_E(\{v_l^i, v_{l+1}^j\})=|W^{(l+1)}_{j, i}|/w_\text{max}$,
where $W^{(l+1)}_{j, i}$ is the weight associated to the edge connecting the vertices $v_l^i$ and $v_{l+1}^j$, as in Figure~\ref{fig:neural_network_explained}, and $ w_\text{max}=\max_{l, i, j}|W^{l}_{i, j}|$ is the maximum weight in absolute value among all the weights of the network.

To compare different layers from the network, \citeauthor{rieck2018neural}\ proposed to normalize neural persistence by dividing it by an upper bound of the neural persistence,
\begin{equation*}
    \text{NP}_l^+(\mathcal{N})= w_{\text{max}}^{-1}\left(\max_{i, j}\left|W^{(l+1)}_{i,j}\right| - \min_{i, j}\left|W^{(l+1)}_{i,j}\right|\right)\left(N_l-1\right)^{1/p},
\end{equation*} 
obtaining the normalized neural persistence $\widetilde{\text{NP}}_l(\mathcal{N})$ for each layer $l$. Averaging normalized neural persistences over all the layers of the network, a global topological complexity measure $\overline{\text{NP}}(\mathcal{N})$ is obtained.

For simple FCFNN networks trained on MNIST, neural persistence was able to distinguish clearly between properly and badly trained networks, for which the values of the neural persistence of properly trained ones were consistently higher than for their badly trained counterparts. Furthermore, it was observed that different regularization techniques augmented the mean neural persistence with respect to the values of regular trained networks, suggesting that for a fixed architecture, the higher the neural persistence, the better the generalization capabilities of the network. This was exploited as an early stopping criterion for training neural networks, where the training process is completed when the mean neural persistence $\overline{\text{NP}}(\mathcal{N})$ stops increasing significantly. This early stopping criterion was found to be competitive with other early stopping criteria but without the need for a validation dataset, whose use is not always possible due to data scarcity.

\citeauthor{caveats_of_neural_persistence_in_deep_neural_networks}\ deepened more into the properties of neural persistence. On the one hand, they discovered tighter bounds than those originally presented by~\cite{rieck2018neural} for the value of neural persistence. On the other hand, they also observed that there is a close relationship between the variance of the learned weights of deep learning models and the neural persistence, questioning the value of the latter with respect to this simpler measure to study the properties of the neural network, arguing that this variance may be similarly useful for the applications showcased by~\citeauthor{rieck2018neural}

\category{Activations whose dissimilarities depend on the weights}

\noindent Both weights and activations can be used together to study the topology of neural networks. In particular, they can be used to inspect differences in the internal workings of neural networks for different input values. Given an input sample $x$ and a neural network $\mathcal{N}$, \cite{characterizing_the_shape_of_activation_space} proposed to analyze the zeroth persistence module $\mathbb V_0(\text{VR}^1(V(G_x), d_\downarrow^0))$ for $G_x$ the undirected weighted graph induced by the directed neural network graph $G(\mathcal{N})$ with weights given by the formula $w_E(\{v_l^i, v_{l+1}^j\})= |W_{j, i}^{(l+1)}\phi_\mathcal{N}^{(l)}(x)_i|$. The weight function captures how activations are distributed through the neural network, as in the previous works. However, in this case, activations are weighted by the weights of the neural network, which do not necessarily depend on the example $x$, and which have an effect of normalization on the influence of the input. Each persistence module is intended to capture information of the network under the influence of input $x$. In this way, the differences between persistence modules for the same network and different inputs can be used to gain insight into the dynamics and representations used by the neural network to compute its function. 

To measure the differences between persistence modules coming from the same network $\mathcal{N}$ and different inputs $x$, $x'$ quantitatively, \citeauthor{characterizing_the_shape_of_activation_space}\ proposed to calculate a dissimilarity based on the differences between the cycle representatives of the points of the two persistence diagrams $D_x= D(\mathbb V_0(\text{VR}^1(V(G_x), d_\downarrow^0)))$ and $D_{x'}= D(\mathbb V_0(\text{VR}^1(V(G_{x'}), d_\downarrow^0)))$, respectively. More precisely, let $\left\{\alpha_i\right\}_{i=1}^{\left|D_x\right|}$ and $\left\{\beta_j\right\}_{j=1}^{\left|D_{x'}\right|}$ be the representatives of the points of $D_x$ and $D_{x'}$, respectively. Each representative cycle $\alpha_i$ and $\beta_j$ is associated with a subgraph of the graph $G_x$ and $G_{x'}$, respectively, denoted by $T_i$ and $T'_j$. Build two vectors $v_x$ and $v_x'$ with as many components as edges there are in the union of graphs $\left(\cup_{i=1}^{\left|D_x\right|}T_i\right)\cup\left(\cup_{j=1}^{\left|D_{x'}\right|}T'_j\right)$, respectively, where $v_x$ and $v_y$ have one in the components corresponding to the edges that come from the union of graphs $\left(\cup_{i=1}^{\left|D_x\right|}T_i\right)$ and $\left(\cup_{j=1}^{\left|D_{x'}\right|}T'_j\right)$, respectively, and a zero otherwise. Then, the dissimilarity between the persistence modules generated from $x$ and $x'$ is defined as a weighed version of the Hamming distance between $v_x$ and $v_{x'}$ using the persistence of sthe cycle representatives associated with the edges corresponding to the different components of the vectors as weights.

The utility and relevance of such persistence modules and their dissimilarity to the study of neural networks was proven in several ways for simple scenarios involving the MNIST, FashionMNIST, and CIFAR-10 datasets and three simple convolutional architectures, including a variant of AlexNet~\citep{alexnet_neural_network} for the CIFAR-10 dataset.  
For example, persistence modules showed excellent classification performance as input to support vector machines (SVMs) using kernel dissimilarity. In practice, the inputs were transformed into their persistence modules for a fixed trained network~$\mathcal{N}$, from which the classification was performed. This approach was tested for some selected trained neural networks, for which the trained SVMs were accurate, even for some adversarial examples. 
Furthermore, dissimilarity between persistence modules was found to resemble distances between the original images in the input space. This suggests that much of the information used by the neural network to classify the examples is contained in the topology as extracted by the persistence modules and that this information is different enough to understand the differences between the behavior of the neural network with respect to different classes. 

The aforementioned persistence modules were further studied to detect adversarial examples by~\cite{goibert2022an}. The hypothesis is that only a small set of edges within the neural networks are used for inference of non-adversarial inputs and that, for adversarial examples, the number of edges used for inference is larger. The idea behind the hypothesis is that adversarial examples attack input-output edge paths with underused edges of the neural network to, with imperceptible modifications of the inputs, completely change the output. Ideally, these changes in the activations of the neurons included in the underused paths make a change in the structure of the activations of the neural network and thus in the persistence modules. To further discern these topological changes, \citeauthor{goibert2022an}\ only add edges to the graph $G_x$ that are underoptimized, that is, edges with \textit{low} weights in absolute value. 

The previous hypothesis is reflected in the number of points of zeroth persistence diagrams points coming from persistence modules of adversarial and non-adversarial examples, where adversarial examples had more points on average than their non-adversarial counterparts in~\citeauthor{goibert2022an} experiments. Furthermore, they successfully detected adversarial examples by training a support vector machine using the sliced Wasserstein kernel~\citep{sliced_wasserstein_kernel} between the persistence diagrams induced by these persistence modules for a variety of neural network types (LeNet, ResNet), datasets (MNIST, Fashion-MNIST, SVHN, CIFAR-10) and adversarial attacks.

Topological data analysis has also been used to analyze the use of neural networks in reinforcement learning (RL) tasks. \cite{topological_dynamics_of_neural_networks_during_rl} studied the evolution of Betti numbers given by homology groups of complexes induced by graphs $G_x^{r, d}$ coming from RL neural networks during a \textit{time} period either in inference or training. Specifically, homology groups for the directed flag~\citep{flagser_algorithm} and Vietoris--Rips complexes were calculated from adjacency graphs $G_x^{r, d}$ derived from the weighted activation graph $G_x$ proposed by~\citeauthor{characterizing_the_shape_of_activation_space}, where $x$ was the input of the RL agent at a specific time. The adjacency graph $G_x^{r, d}$ was built by taking the edges of $G_x$ with weights higher than or equal to $r$ and adding an edge between two vertices $v_l^i$ and $v_l^j$ if there existed a third vertex $v_{l-1}^k$ such that $\left|(W_{i,k}^{(l)} - W_{j,k}^{(l)})\phi^{(l-1)}_\mathcal{N}(x)_k\right|<d$. By including these new edges, connections are added between neurons whose activations are highly related. 

For inference experiments, Betti numbers up to dimension three for each time step were computed and \text{smoothed} using a moving average window of size four. Transitions between actions of the agent and evolution of the Betti numbers of dimension three seemed to be correlated, while for the other dimensions this correlation was not conclusive. For the training experiments, complex environments seemed to induce the development of higher-order Betti numbers during training. Also, with a similar number of neurons, the higher the Betti numbers in the last steps of training, the better the model worked. 

For the training experiments, \citeauthor{topological_dynamics_of_neural_networks_during_rl}\ studied a matrix $H$ with training steps as columns and neurons as rows, where each coefficient $H_{v, t}$ of the matrix was derived from cocycle representatives of the one-dimensional cohomology groups of the Vietoris--Rips complexes at time $t$. Specifically, $H_{v,t}$ was the highest cardinality of the set of edges on the support of a set of cocycles $\mathcal{C}^t_v$, where $\mathcal{C}^t_v$ contained cocycles whose support had edges containing the vertex $v$ for each node $v$ at time $t$. Hence,
\begin{equation*}
    H_{v,t}= \max_{c\in \mathcal{C}^t_v}\left|\text{supp}(c)\cap E(G_x^{r, d})\right|.
\end{equation*}
The matrix $H$ was seen to have higher values for neurons on the latest layers consistently for all the training steps, suggesting that deeper neurons have more relevance in the topological structure extracted from the graphs $G_x^{r, d}$. However, more experiments are needed to validate these results, as experiments were performed only for basic FCFNNs.

A combination of the methods proposed in~\cite{rieck2018neural} and in~\cite{characterizing_the_shape_of_activation_space} is used to define a new prediction reliability score for neural networks in classification problems called \textit{topological uncertainty}. Given a neural network $\mathcal{N}$ and a sample $x_\text{in}$ for which we want to compute this score, the topological uncertainty of $x_\text{in}$ is computed from the set of persistence diagrams 
\begin{equation*}
   D^l_x= D^w(\mathbb V_0(\text{VR}^1(V(G_x^l) ,d_V^0))),
\end{equation*}
for all $l\in[L]$, $x\in\{x_\text{in}\}\cup \left\{x:(x,y)\in \mathcal{D}_\text{train}\right\}$, where $L$ is the number of non-input layers of $\mathcal{N}$, $\mathcal{D}_\text{train}$ is the training dataset, and $G_x^l$ is the subgraph of $G_x$ as in the previous work by~\cite{characterizing_the_shape_of_activation_space} induced by the vertices of the layer $l$ and $l-1$ of $\mathcal{N}$ and all the possible edges connecting them in $G_x$ whose edges are weighed as in $G_x$ and whose vertices are weighed as $-\infty$. We do this to always have a fixed number of points equal to the number of vertices of $G_x^l$ for any possible weight assignment to the edges and to have all birth values equal. It can be proved that, in this way, there is a bijection between the finite deaths of the persistence diagrams $D^l_x$ and the multiset of weights of the maximum spanning tree of $G_x^l$, which we denote by $\mathfrak W_x^l$. These persistence diagrams are meant to capture information similar to that computed in~\cite{characterizing_the_shape_of_activation_space}, but more fine-grained, since they are computed for each pair of adjacent layers. 

Recall that $\left|\mathfrak W_x^l\right| = \left|\mathfrak W_{x'}^l\right|$ for any pair of inputs $x$ and $x'$. Given a multiset of weights $\mathfrak W_x^l$, let us order them in descending order and denote them by $\mathfrak W_x^l= \{w_{x, 1}^l, \hdots, w_{x, \left|W_x^l\right|}^l\}$. These weights induce a probability distribution on each of the subgraphs $G_x^l$ given by 
$\mu_x^l= \frac{1}{n}\sum_{i=1}^{n}\delta_{w_i}$, 
where $n=\left|\mathfrak W_x^l\right|$ and
$\delta_{w_i}$ denotes the Dirac measure at $w_i\in\mathbb R$. For the same layer, these probabilities distributions can be combined to obtain an \textit{average topological distribution} across several samples. This average between an arbitrary number of points $x_1,\hdots, x_m$ is given by
\begin{equation*}
    \bar{\mu}^l= \frac{1}{\left|\mathfrak W_x^l\right|}\sum_{i=1}^{\left|\mathfrak W_x^l\right|}\delta_{\bar{w}_i}, \qquad \bar{w}_i = \frac{1}{m}\sum_{j=1}^m w_{x_j, i}^l.
\end{equation*}
Therefore, given the new sample $x_\text{in}$, the topological uncertainty measures the average difference between the average topological distributions $\bar{\mu}^l$ for each layer $l$ calculated for the subset of the training dataset $\mathcal{D}_\text{train}$ with label equal to the predicted label for $x_\text{in}$ with respect to the topological distributions $\mu^l_{x_\text{in}}$ of the input sample $x_\text{in}$, that is,
\begin{equation*}
    \text{TU}_x(\mathcal{N}) = \frac{1}{L}\sum_{l=1}^L\text{d}(\mu_{x_\text{in}}^l, \bar{\mu}^l_{\mathcal{D}_x}),\quad \mathcal{D}_x= \{x:(x,y)\in\mathcal{D}_\text{train}\text{ with }\pi\circ\phi_\mathcal{N}(x_\text{in})=y\},
\end{equation*}
where $\bar{\mu}^l_{\mathcal{D}_x}$ is the average probability distribution over the set $\mathcal{D}_x$ and where $\text{d}(\mu_1, \mu_2)= \frac{1}{\left|\mathfrak W_1\right|}\sum_{i=1}^{\left|\mathfrak W_1\right|}\left|w^1_i - w^2_i\right|$ is a dissimilarity function between distributions coming from sets of weights $\mathfrak{W}_1=\{w^1_1,\hdots, w^1_{\left|\mathfrak W_1\right|}\}$ and $\mathfrak{W}_2=\{w^2_1,\hdots, w^2_{\left|\mathfrak W_1\right|}\}$ with the same number of points ordered as described. 

The lower the value of the topological uncertainty measure, the more reliable the prediction for $x_\text{in}$ as its internal behavior is more similar to the behaviour of the network for the examples in the training dataset with the same label. This measure was used to successfully perform model selection from a bank of models trained on MNIST and Fashion-MNIST, where the model with the lowest average topological uncertainty for the new dataset is selected,  and out-of-distribution and shifted examples detection for several basic networks and datasets, including MUTAG~\citep{mutag_dataset}, COX2, and MNIST datasets, considering only some set of layers (non-convolutional layers) to compute the topological uncertainty in some cases. 

\category{Generic spaces}

\noindent So far we have seen many methods to compare differences between the topology, in a broad sense, of different neural network \textit{representations}. Most of these methods are simply based on distances on either persistence diagrams or on some constructions coming from the persistence modules. A more direct approach is taken by~\cite{representation_topology_divergence}, in which they propose a method to compare two Vietoris--Rips filtrations for the same set of points $V$ and different distances $d_1$, $d_2$. The idea behind the method is to compare, given a specific threshold $t$, how the connected components of $\text{VR}_t(V,d_1)$ and $\text{VR}_t(V, d_2)$ are merged in $\text{VR}_t(V, d_\text{min})$ where $d_\text{min}(x,y)=\min(d_1(x,y), d_2(x,y))$. In particular, they count how many connected components are merged at each threshold for all the possible thresholds, and derive a measure of dissimilarity from such countings. This measure of dissimilarity is called \textit{representation topology divergence}  and is defined as the average of two total persistences divided by two, denoted by $\text{RTD}_1(d_1, d_2)$ and $\text{RTD}_1(d_2, d_1)$, calculated from the one-dimensional persistence diagrams computed from the Vietoris--Rips filtrations of the point clouds $(V_{1, 2}, d_{1, 2})$ and $(V_{2, 1}, d_{2,1})$ with vertices $V_{i, j} = \left\{v_a\right\}_{a=1}^{\left|V\right|} \cup \left\{v'_a\right\}_{a=1}^{\left|V\right|} \cup \{O\}$ and distances given by
\begin{equation*}
    \begin{split}
        d_{i, j}(v'_a, v'_b) & = \min(d_i(v_a, v_b), d_j(v_a, v_b)),\\[0.1cm]
        d_{i, j}(v_a, v'_b) & = d_{i, j}(v_a, v_b) = d_i(v_a, v_b), \\[0.1cm]
        d_{i, j}(v_a, v'_a) & = d_{i, j}(O, v_a) = 0, \\[0.1cm]
        d_{i, j}(v_b, v'_a) & = d_{i,j}(O, v'_a) = +\infty,
    \end{split}
\end{equation*}
for $i\in[2]$, $j\in[2]\smallsetminus\{i\}$, where $v'_a$ is the node $v_a$ duplicated in the point cloud and $O$ is an abstract point that is useful to capture the differences between $\text{VR}_t(V, d_i)$, $\text{VR}_t(V, d_j)$  and $\text{VR}_t(V, d_\text{min})$. Intuitively, the $k$-dimensional persistence diagram from the Vietoris--Rips filtration of these point clouds records the $k$-dimensional topological features that are born in $\text{VR}_t(V, d_\text{min})$ but not yet in $\text{VR}_t(V, d_i)$, and the $(k-1)$-dimensional topological features that are dead in $\text{VR}_t(V, d_\text{min})$ but are not yet dead for $\text{VR}_t(V, d_i)$.

The representation topology divergence has been used to analyze many aspects of neural networks. For example, \citeauthor{representation_topology_divergence}\ used it to analyze $400$-dimensional embeddings of $10{,}000$ words for $90$ randomly selected architectures from the NAS-Bench-NLP~\citep{klyuchnikov2020nasbenchnlp} and several properties of neural networks, among others. For neural networks, they trained a VGG-11 and a ResNet-20 convolutional networks on CIFAR-10 and CIFAR-100 and compared the evolution of the activations of the convolutional layers to the activations of the final trained network. They observed that the representation topology divergence between activations decreased as the number of epochs trained increased, capturing the convergence to the final state of the network. They also compared the representations of the images given by different layers of the network for both original and shifted examples, observing differences between the representations of the samples in different layers and being able to detect the shifted examples. Furthermore, they showed that the representation topology divergence could be a good indicator of the diversity of the models in an ensemble method used in an appropriate way. Finally, they found that~the representation topology divergence comparing the trained models correlated well with the disagreement of the predictions, although more experiments are needed to understand this relationship.

\subsection{Training dynamics and loss functions}\label{scn:training_dyanmics_loss_functions}
In this section, we review only three articles that focus on studying the properties of the training process. The first one deals with the loss function used to train the neural network. The other two are focused on the evolution of the weights during the training process and how the \textit{fractal dimension} of weight trajectories are related to the generalization capacity of neural networks.

\category{Loss functions}

\noindent One of the fundamental problems in deep learning is, given a learning problem, an architecture $a$, a loss function $\mathcal{L}$, and a training algorithm $\mathcal{A}$, to determine if the training algorithm $\mathcal{A}$ is capable of finding a neural network $\mathcal{N}$ with architecture $a$ that minimizes the empirical risk $\widehat{\mathcal{R}}_{\mathcal{D}_\text{train}}$ for the learning problem. Usual training algorithms perform a gradient descent, following a path in the space of parameters of the neural network. This path is then heavily affected by the connectivity of the loss graph and by the presence of \textit{local valleys}, in which the gradient descent algorithm can get stuck.

\cite{on_connected_sublevel_sets} studied a generalization of this problem for general FCFNNs depending on their activation functions, their graph structure (depth and width of layers), and the \textit{shape} of the training dataset. Specifically, he studied the number of connected components and the existence of local valleys on the graph of an optimization target $\mathfrak{L}(\theta)= f(\phi_{\mathcal{N}_\theta}(x_1),\hdots,\phi_{\mathcal{N}_\theta}(x_m))$, where $f$ is a convex function, and $\mathcal{D}_\text{train}=\{(x_i, y_i)\}_{i=1}^m$. These different optimization targets allow training algorithms to minimize any convex function with respect to the parameters of neural networks that could be useful to obtain better parameters for the network, not restricting the minimization to empirical risks. Although this is a more complex scenario than the one presented in Section~\ref{scn:deep_learning_fundamentals}, empirical risk functions for many loss functions, such as categorical cross entropy, are examples of these types of optimization targets. For the discussion of this article, we assume that the activation functions for all layers except for the last one are the same, and that the last-layer activation function is simply the identity. 

The first insight on the optimization target graph is that, for strictly monotonic activation functions $\varphi$ with $\text{Im}(\varphi)=\mathbb R$, FCFNNs with widths strictly decreasing layer by layer, that is, $N_i > N_{i+1}$ for $i\in[L-1]$ and with at least two non-input layers, i.e., $L\geq 2$, and a linearly independent training dataset $\mathcal{D}_\text{train}$, every sublevel set of $\mathfrak{L}$, that is, the sets $\mathfrak{L}^{-1}(-\infty, \alpha]$ for $\alpha\in\mathbb R$, is connected and also every non-empty connected component of every level set $\mathfrak{L}^{-1}(\alpha)$ is unbounded. Connectedness of sublevel sets implies a well-behaved optimization target function, and unboundedness of level sets implies that there are no local valleys in the optimization target graph, understanding by a local valley a non-empty connected component of some strict sublevel set $\mathfrak{L}^{-1}(-\infty, \alpha)$.

A bad local valley is a local valley in which the target $\mathfrak{L}$ cannot be arbitrarily close to the lower bound of the convex function inducing the target function, which is also a lower bound of $\mathfrak{L}$ but does not depend on any concrete neural network. Bad local valleys are harmful to the optimization problem because if the training process enters them when optimizing the target function, the parameters obtained at the end of the learning process are provably not optimal. \citeauthor{on_connected_sublevel_sets} proved that, for activation functions as before that also satisfy that there are no nonzero coefficients $(\lambda_i, a_i)_{i=1}^p$ with $a_i\neq a_j$ for all $i\neq j$ such that $\varphi(x) = \sum_{i=1}^p\lambda_i\varphi(x-a_i)$ for all $x\in\mathbb R$, FCFNNs with a layer $l\in[L-1]$ satisfying $N_l > \left|\mathcal{D}_\text{train}\right|$ and $N_i > N_{i+1}$ for $i\in \{l+1,\hdots, L\}$ induce target functions $\mathfrak{L}$ with no bad local valleys and, if $l\leq L-2$, then every local valley of $\mathfrak{L}$ is unbounded. This implies that from any initial parameter, there is a continuous path on which the loss function is nonincreasing from it to a point that is arbitrarily close to the infimum of the loss. 

As we have seen so far, widths of neural network layers play a significant role in the configuration of the optimization target graph. The first hidden layer always plays an important role in this configuration, since it determines the first transformation of the data into some space from which features from the inputs are extracted. Assuming that the activation functions of the network satisfy the two previous assumptions, and assuming that $N_1 > 2\left|\mathcal{D}_\text{train}\right|$ and that $N_i > N_{i+1}$ for $i\in\left\{2,\hdots, L-1\right\}$, \citeauthor{on_connected_sublevel_sets} proved that every sublevel set of $\mathfrak{L}$ is connected and also that every connected component of every level set of $\mathfrak{L}$ is unbounded. This is a stronger result than the previous one, as it implies that not only there are no bad local valleys but also there is a unique global valley. 

Many current activation functions, such as the leaky ReLU, satisfy the previous assumptions. However, the usual ReLU does not and further assumptions are needed to have a good behavior of the target graph. Letting $N_\text{min}=\min_{i\in [L-1]}N_i$, for FCFNNs with activation functions $\varphi$ only satisfying the last assumption implying coefficients $(\lambda_i, a_i)_{i=1}^p$, if $N_\text{min} > \left|\mathcal{D}_\text{train}\right|$, then $\mathfrak{L}$ has no bad local valleys and, if $N_\text{min} > 2\left|\mathcal{D}_\text{train}\right|$, then every sublevel set of $\mathfrak{L}$ is connected. 

Some of the previous assumptions are too strong to be satisfied in many practical scenarios. However, these results serve as a basis for a better understanding of the shape of the target function. A similar approach using topological data analysis to capture the connectivity and shape of optimization target graphs for modern neural networks could be a promising research line. This could help verify if the claims given in these cases can be extrapolated to more general scenarios.

\category{Fractal dimension of weight trajectories}

\noindent The last approach that we discuss in this survey is related to the evolution of parameters during the training process. In this case, training algorithms $\mathcal{A}$ are assumed to be continuous, which means that the parameters evolve over a time period $[0, T]\subseteq \mathbb R$ for which each instant of time$t\in[0, T]$ has an associated weight value $\theta_t$. Although this is not the case for real scenarios, since computers work in the discrete domain, there are good continuous approximations of classical \textit{discrete} optimization algorithms such as gradient descent that allow us to study these discrete processes with properties of the continuous approximations. 

\cite{hausdorff_dimension_heavy_tails} discovered that the generalization capacity of neural networks was connected with the heavy-tailed behavior of weight trajectories $\Theta_\mathcal{A}=\left\{\theta_t: t\in[0, T]\right\}$ generated during training, in particular, with its upper box dimension, a dimensionality measure for fractals. A~difficulty with this link is that many strong assumptions about the training algorithm and the space of weights were needed to formally prove the connection. 

\cite{intrinsic_dimension} relaxed these assumptions by cleverly using a previous result that connected this fractal dimension with persistent homology~\citep{mst_ph_1, mst_ph_2}. This result says that, for $\Theta\subseteq R^d$ a bounded set, we have
\begin{equation*}
    \text{dim}_\text{PH}\Theta = \text{dim}_{\text{PH}}^0\Theta=\text{dim}_\text{Box}\Theta,
\end{equation*}
where $\text{dim}_\text{Box}$ is the upper-box fractal dimension, and $\text{dim}_{\text{PH}}^k$ is the persistent homology dimension given by
\begin{equation*}
\begin{split}        \text{dim}_{\text{PH}}^k\Theta &= \inf \left\{\alpha: E_\alpha^k(\Theta_{<\infty}) < C: \exists C>0\text{ for all finite } \Theta_{<\infty}\subseteq \Theta\right\}, \\
E_\alpha^k(P) &= \sum_{(b,d)\in D}\left|d-b\right|^\alpha, \qquad D=D(\mathbb V_k(\text{VR}(P, \lVert \cdot \rVert_2))),
\end{split}
\end{equation*}
that is, the infimum of all the exponents for which $E_\alpha^k$ is uniformly bounded for all finite subsets $\Theta_\text{inf}$.  

In particular, \citeauthor{intrinsic_dimension}\ proved that, given any compact set of (random) possible weights $\Theta$, for example, the weight training trajectories $\Theta_\mathcal{A}$, $\mathcal{L}(f, x_i, y_i)$, and a loss function $\mathcal{L}$ bounded by $B$ and $K$-Lipschitz continuous on the set of possible parameters $\theta$ of a fixed neural network architecture, then, for a sufficiently large size of the training dataset $m$, the following holds:
\begin{equation*}
\sup_{\theta\in\Theta}\left|\widehat{\mathcal{R}}_{\mathcal{D}_\text{train}}(\phi_{\mathcal{N}_\theta}) - \mathcal{R}(\phi_{\mathcal{N}_\theta})\right| \leq 2B\sqrt{\frac{(\text{dim}_\text{PH}\Theta + 1)\log^2(mL^2)}{m} + \frac{\log(7M/\gamma)}{m}}
\end{equation*}
with probability at least $1-\gamma$ over a training dataset $\mathcal D_\text{train}$ with $m$ elements sampled i.i.d.\ from the data distribution, where $\mathcal{N}_\theta$ denotes the neural network $\mathcal{N}$ defined with the fixed architecture and parameters $\theta$ and $M$ denotes a constant depending on some technical assumptions about the objects involved in the bound.

The previous bound uses persistent homology dimension, which cannot be computed exactly by a computer program. \citeauthor{intrinsic_dimension} propose an algorithm to estimate this quantity for finite sets of weights. 
Estimations of persistent homology dimension were found to be significantly correlated with the generalization gap calculated as the difference between the accuracy in train and the accuracy in test, indicating that lower persistent homology dimensions were associated with better generalization capacities of networks for a wide variety of networks, including simple FCFNN models and AlexNet, trained on MNIST, CIFAR-10, and CIFAR-100 datasets.

Finally, the differentiability theory for persistence diagrams allows one to minimize these estimations as a regularization method, expecting to obtain better generalizations for regularized models. This is exactly what happened for a LeNet-5~\citep{leNet} architecture trained on CIFAR-10 in the experiments performed by~\citeauthor{intrinsic_dimension}, especially for training procedures that failed to converge to a good set of parameters only by themselves.

The previous bounds were improved in a follow-up paper by~\cite{generalization_bounds_using_data_dependent} in which the persistent homology dimension was used again to compute fractal dimensions. As the methods employed to improve the bounds are not related with topological data analysis, we do not analyze it in this survey, although the results are very insightful and interesting in their own right.

\section{Challenges, future directions, and conclusions}\label{scn:challenges_future_directions_conclusions}

In this survey we have seen many examples of how topological data analysis, and particularly (persistent) homology and Mapper, can be applied to study the properties of neural networks. In Section~\ref{scn:structure_of_the_neural_network}, we saw how to compute two homology groups that yielded different information about a neural network only by taking its graph without weights. However, we noticed that the ranks of 
those homology groups were invariant to many important structural properties of the neural network, such as the order of the layers. Also, their values were simply a combination of the width and the depth of the graph, not capturing much information about the network. To obtain better insight into the properties of neural networks, more fine-grained homology theories for directed graphs are needed. A~promising line of work would be to study the persistent path homology~\citep{persistent_path_homology} of graphs augmented with the weights associated to each edge, as in Section~\ref{scn:internal_representations_and_activations}.

In Section~\ref{scn:input_and_output_spaces}, we saw how topology was useful in recovering the \textit{topology} of decision regions and boundaries, with applications especially in model selection. Furthermore, we saw that GTDA, an evolution of the Mapper algorithm for graphs as input, was successful in analyzing patterns in output spaces of neural networks, leading to a better understanding of misclassification errors in several datasets.

Thanks to the differentiability theory of persistence diagrams, we saw how to \textit{improve} decision regions by \textit{simplifying} them. In addition, we discussed how the topology was useful for capturing the quality ---measured in different ways--- of generative models. Particularly interesting is the study of the disentanglement of generative models, since being able to decouple the different sources of variation could lead to a better control and quality of the networks' outputs. We find that refining the article by~\cite{zhou2021evaluating} and using its measures to regularize generative neural networks could be an interesting approach to improving neural network design using TDA.

In Section~\ref{scn:internal_representations_and_activations} we analyzed the methods studying the neural network parameters and activations. We split the section into the articles using Mapper and (persistent) homology, because of the differences in their approaches. In the first case, we found both studies of the parameters and of the activations. In the first case, we saw how~\citeauthor{topological_approaches_to_deep_learning} found very interesting connections between the topologies of the space of natural images and the space of weights of some layers in convolutional networks, among many other interesting topology configurations of the convolutional weights. Also, we saw how~\citeauthor{topology_of_learning_in_feedforward_neural_networks} studied the Mapper graphs of the evolution of the weights during training, finding a very interesting pattern in the weight distribution of one of the architectures tried in his experiments: the weights seemed to be distributed in a \textit{surface!} A closer look at the topology of the weight evolution for a higher variety of architectures must be taken to realize if the weights share common patterns in similar problems. 
This could potentially enable leveraging this structure, as proposed in~\cite{topological_approaches_to_deep_learning}, for instance. For activations, we found that the general approach proposed by TopoAct~\citep{topoact} was extremely useful for understanding the internal behavior of neural networks, with the many ramifications presented in the survey.

For (persistent) homology papers, we could broadly classify the approaches depending on either
the set of neurons/edges analyzed (layer by layer versus all the neurons), or the dissimilarity strategy (weights versus distances between activations or versus a combination of weights and activations).

In many cases, there was a connection between the different topological summaries extracted from Vietoris--Rips persistence diagrams and the properties of neural networks, especially with their generalization capacities. However, ~\citeauthor{caveats_of_neural_persistence_in_deep_neural_networks}\
showed that most of the information extracted by one of the most influential articles in this section could be captured by taking simpler, non-topological summaries of the neural network. Although this does not mean that TDA does not provide unique information about the internal workings of the neural networks, we think that more ablation studies must be performed in the studies claiming the utility of TDA for the analysis of neural networks, as almost all the work performed in this area is experimental, and, usually, there is no theory supporting the hypotheses connecting topological summaries and neural network properties. 

Another interesting discussion in this subsection is the opposition of views on the evolution of the topology of the data through the different layers. In most works, such as the one by~\cite{topology_of_deep_neural_networks}, it is stated and verified that the topology of the data is \textit{simplified} by the action of the layers. However, the work by~\cite{activation_landscapes_as_a_topo_summary} arrived to the opposite conclusion. This contradiction may have occurred due to the simplicity of the experiments performed in all of the experiments and supports the necessity of performing more diverse and complete experiments.

The previous point leads to one of the fundamental drawbacks of the section: most of the experiments are performed on classical CNN and FCFNN architectures and do not say anything about more modern architectures, like transformers. To be useful, TDA must be applied to state-of-the-art architectures, not only as a proof of concept on \textit{simple} neural networks. This is a fundamental step for gaining the trust of the non-TDA deep learning community in TDA methods. Also, since many people working on this area are mathematicians, it would be desirable that TDA methods be accompanied by theoretical results, as in~\cite{intrinsic_dimension}.

In Section~\ref{scn:training_dyanmics_loss_functions}, we saw how TDA was used to study the loss function and training weights. In this case, we observed that persistence diagrams, which come originally from topology, were used to compute fractal dimensions, that belong to the geometry realm. This is an exciting result because it means that persistent homology can be used not only to infer the topology of the data, but also to infer geometrical properties. This is further studied by~\cite{andreeva2023metric}, and we think that seeing persistent homology as a tool that also extracts local information from the data could be useful in performing new work to analyze the structures of neural networks.

Probably the most critical drawback of topological data analysis for neural networks is the high computational complexity in time and memory of computing invariants, like persistence diagrams, of persistence modules. For dimension zero, algorithms based on the minimum spanning tree~\citep{minimum_spanning_tree_acker} or the single linkage algorithm~\citep{slink_algorithm}, have time complexities $\mathcal{O}(e\cdot \alpha(e,n))$ and $\mathcal{O}(n^2)$, respectively, where $\alpha$ is the very slow growing functional inverse of the Ackermann function~\citep{efficient_set_union_algorithm} and $e$ is the number of edges of the simplicial complex, whose cardinality is $\mathcal{O}(n^2)$. Also, for single-linkage clustering, the memory complexity is linear, that is, $\mathcal{O}(n)$. This makes these algorithms suitable for some of the applications for which the number of points is relatively small, such as the analysis of weights or activations layer by layer. However, for bigger point clouds, such as in the problems in which we use the whole set of neurons or weights, these algorithms become infeasible for modern and big neural networks. The situation is worse for dimensions greater than or equal to one, where typical persistence diagram computation algorithms have a time and memory complexity of $\mathcal{O}(n^w)$ and $\mathcal{O}(n^2)$, respectively, where $w$ is the matrix multiplication exponent (currently $w<2.4$) and $n$ is the number of simplices generated through the filtration of simplicial complexes~\citep{intrinsic_dimension}. 

Computational problems motivate the development of more suitable algorithms to compute persistence diagrams or alternative invariants that also capture information of neural networks. Although powerful tools for computing Vietoris--Rips persistence diagrams are available, such as Ripser~\citep{ripser_paper}, Ripser++~\citep{zhang2020gpu}, Gudhi~\citep{gudhi}, SLINK~\citep{slink_algorithm}, algorithms that parallelize computations are of special interest due to the increasing availability of powerful hardware and software methods to perform distributed computing. Also, approximating known filtrations by means of new filtrations, as in~\citep{linear_size_approximations}, or by means of machine learning models, as in~\cite{montufar2020can}.

As we have seen in this survey, there are many different ways to produce meaningful filtrations for neural networks in all the domains we reviewed. For this reason, it would be desirable to capture the joint information of some of them at the same time. Also, for some specific problems, like analyzing the evolution of the data distribution through the layers, filtrations by distances alone may not be good enough, as they do not consider important data properties such as the density of points in the sample taken. Luckily, multiparameter persistence~\citep{botnan2023introduction} deals with this kind of problem setting, allowing to extract information from non-linear filtrations, that is, with more than one varying parameter. Unluckily, multiparameter persistence does not have a canonical, easy-to-use and computationally efficient representation like persistence diagrams for one-dimensional persistence. However, there are already works proposing a rich body of useful, and even differentiable, representations for multiparameter persistence; see, for example~\cite{loiseaux2023a, loiseaux2023stable, gril_mp}. We see multiparameter persistence as one of the main lines of work not only for the analysis of deep neural networks, but also for the whole machine learning community, as multiparameter persistence provides sharper ways to extract information from data. However, we are far from having usable representations for real neural network use cases due to the huge quantity of points in real datasets and neurons in architectures. For this reason, further fundamental research on the topic is needed before we can grasp its advantages in the deep learning community. 
Further work in this direction could be based on three basic pillars:
\begin{enumerate*}
    \item Ease of use of 
    multiparameter representations that can be stored in any conventional computer, since representations must be intuitive for the average machine learning researcher;
    \item Efficiency of computation;
    \item Differentiability of representations with respect to the point clouds, to allow regularization in learning tasks.
\end{enumerate*}

In conclusion, we have seen that the use of TDA applied to studying neural networks is an exciting path that can be useful in many scenarios. TDA, although computationally expensive, is a tool that has been shown to be connected with many interesting properties of neural networks such as generalization in classification problems or entanglement and latent space quality in generative models. 
Moreover, we have seen that there are many ways in which TDA can be used in applied scenarios, converting TDA into an essential tool for deep learning practitioners.


\acks{This work was supported by the Ministry of Science and Innovation of Spain through projects PID2019-105093GB-I00, PID2020-117971GB-C22, and PID2022-136436NB-I00; by the Ministry of Universities of Spain through the FPU contract FPU21/00968, and by ICREA under the ICREA Academia programme.}


\vskip 0.2in

\bibliography{refs}


\newpage

\appendix

\section{Peer-reviewed articles discussed in this survey}

\input{assets/papers_table.tbl}


\newpage

\section{The Mapper algorithm}

\begin{algorithm}[ht]
\caption{Mapper algorithm}\label{alg:mapper_basic}
\begin{algorithmic}[1] 
\REQUIRE $\mathcal{D}$ with $\left|D\right|=m$, filter function $f\colon \mathcal{D}\to\mathbb R^d$, finite cover $\mathcal{U}=\left\{\mathcal{U}_i\right\}_{i\in I}$ of $\text{Im}(f)\subseteq \mathbb R^d$, clustering algorithm $\mathcal{C}$.
\ENSURE Simplicial complex $S_\mathcal{D}$.
\STATE $S_\mathcal{D}\leftarrow \emptyset$
\STATE $\mathcal{D}_i\leftarrow f^{-1}\left(\mathcal{U}_i\right)$ for all $i\in I$
\FORALL{$i \in I$}
    \STATE $\{C_i^1,\hdots, C_i^{k_i}\}\leftarrow \mathcal{C}\left(\mathcal{D}_i\right)$ \COMMENT{Apply the clustering algorithm to $\mathcal{D}_i$: the output are the clusters}
    \STATE $S_\mathcal{D} \leftarrow S_\mathcal{D}\cup \{C_i^1,\hdots, C_i^{k_i}\}$\COMMENT{Add the clusters found as vertices}
\ENDFOR
\FORALL{$\{C_1,\hdots, C_t\}\in\mathcal{P}\left(\bigcup_{i\in I}\{C_i^1,\hdots, C_i^{k_i}\}\right)$\COMMENT{For all possible subsets of found clusters}}
\IF{$\bigcap_{j=1}^tC_j\neq \emptyset$}
\STATE $S_\mathcal{D} \leftarrow S_\mathcal{D}\cup \left\{\left\{C_1,\hdots ,C_t\right\}\right\}$ \COMMENT{We add the simplex $\left\{C_1,\hdots ,C_t\right\}$}
\ENDIF
\ENDFOR
\RETURN $S_\mathcal{D}$
\end{algorithmic}
\end{algorithm}

\end{document}

%% file: assets/papers_table.tbl
\LTcapwidth=\textwidth\small
\renewcommand{\arraystretch}{1.5}
\newcolumntype{R}{>{\raggedright\arraybackslash}}
\begin{longtable}{Rp{0.37\textwidth}Rp{0.43\textwidth}cc}
\caption{Table containing the published articles reviewed in this survey ordered by year of publication. Each row corresponds to an article. The first column contains the title; the second column contains a one-line summary; the third column contains the categories introduced in Section~\ref{scn:introduction} associated to each article; and the last column contains the applications introduced in Section~\ref{sec:applications_of_tda_in_dl}. The four possible categories are 1.\;Structure of the neural network; 2.\;Input and output spaces; 3.\;Internal representations and activations; 4.\;Training dynamics and loss functions. The seven possible applications are 1.\;Regularization; 2.\;Pruning of neural networks; 3.\;Detection of adversarial, out-of-distribution and shifted examples; 4.\;Detection of trojaned networks; 5.\;Model selection; 6.\;Prediction of accuracy; 7.\;Quality assessment of generative models.}
\label{tab:summary_articles}\\
\toprule
\textbf{Title} & \textbf{Summary} & \textbf{Cat.} & \textbf{Apps.} \\
\midrule
\endfirsthead
\toprule
\textbf{Title} & \textbf{Summary} & \textbf{Cat.} & \textbf{Apps.} \\
\midrule
\endhead
\midrule
\multicolumn{4}{r}{Continued on next page} \\
\midrule
\endfoot
\bottomrule
\endlastfoot
\textit{On the complexity of neural network classifiers: A comparison between shallow and deep architectures} \citep{on_the_complexity_of_neural_network_classifiers_a_comparison_between_shallow_and_deep_architectures} & Bounds on the Betti numbers of the positive decision region generated by binary classification neural networks with Pfaffian activations & (2) & - \\
\textit{Topological approaches to deep learning} \citep{topological_approaches_to_deep_learning} & Topological analysis of the weights of convolutional neural networks and generalization of convolutional neural networks & (3) & - \\
\textit{Topological data analysis of decision boundaries with application to model selection} \citep{tda_of_decision_boundaries_with_application_to_model_selection} & Study and approximation of the topology of network decision boundaries & (2) & (5) \\
\textit{A topological regularizer for classifiers via persistent homology} \citep{a_topological_regularizer_for_classifiers_via_persistent-homology} & Regularization of neural networks by modifying their decision regions using differentiable persistent homology & (2) & (1) \\
\textit{Geometry Score: A method for comparing generative adversarial networks} \citep{geometry_score_comparing_generative_adversarial_networks} & Measurement of GAN quality using persistent homology & (2) & (7) \\
\textit{What does it mean to learn in deep networks? And, how does one detect adversarial attacks?} \citep{whatdoesitmean_corneanu} & Study of generalization in terms of the persistent homology of neuron activations and their correlations  & (3) & (1, 3) \\
\textit{On connected sublevel sets in deep learning} \citep{on_connected_sublevel_sets} & Study of the connectivity, boundedness, and local minima of sublevel sets of convex losses for overparameterised neural networks with piecewise linear activation functions & (4) & - \\
\textit{Neural persistence: A complexity measure for deep neural networks using algebraic topology} \citep{rieck2018neural} & Topological complexity measure for neural networks based on their weights & (3) & (1) \\
\textit{Characterizing the shape of activation space in deep neural networks} \citep{characterizing_the_shape_of_activation_space} & Topological characterization of the neuron activations given a sample & (3) & (3) \\
\textit{Exposition and interpretation of the topology of neural networks} \citep{exposition_and_interpretation_of_the_topology_of_neural_networks} & Topological analysis of the weights of convolutional neural networks and connection to generalization capacity of models & (3) & - \\
\textit{Path homologies of deep feedforward networks} \citep{path_homologies_of_deep_feedforward_networks} & Analysis of path and directed flag homology groups of MLPs' directed graph & (1) & - \\
\textit{Topology of deep neural networks} \citep{topology_of_deep_neural_networks} & Study of the topology of the data through layer transformations & (3) & - \\
\textit{Finding the homology of decision boundaries with active learning} \citep{finding_the_homology_of_decision_boundaries_with_active_learning} & Use of active learning to improve the methods of Topological data analysis of decision boundaries with application to model selection & (2) & (5) \\
\textit{Computing the testing error without a testing set} \citep{computing_test_error_corneanu} & Regression of test accuracy using topological vectorizations of persistence diagrams & (3) & (6) \\
\textit{Topological detection of trojaned neural networks} \citep{topo_detection_trojaned} & Study of trojaned networks in terms of the persistent homology of neuron activations and their correlations & (3) & (4) \\
\textit{Experimental stability analysis of neural networks in classification problems with confidence sets for persistence diagrams} \citep{experimental_stability_analysis} & Study of persistence diagrams of last hidden layer activations and its connection with generalization & (3) & - \\
\textit{PHom-GeM: Persistent homology for generative models} \citep{charlier2019pho} & Comparison between persistence diagrams of real and generated manifolds using generative models & (2) & (7) \\
\textit{TopoAct: Visually exploring the shape of activations in deep learning} \citep{topoact} & Analysis of the Mapper graph of neuron activations for each layer & (3) & - \\
\textit{Deep neural network pruning using persistent homology} \citep{network_pruning_using_PH} & Pruning of neural networks using persistent homology & (3) & (2) \\
\textit{Intrinsic dimension, persistent homology and generalization in neural networks} \citep{intrinsic_dimension} & Generalization error bounds using persistent homology dimension on the training weights & (3, 4) & (1) \\
\textit{Activation landscapes as a topological summary of neural network performance} \citep{activation_landscapes_as_a_topo_summary} & Study of the topology of the data through layer transformations and its connection with training accuracy & (3) & - \\
\textit{Topology of learning in feedforward neural networks} \citep{topology_of_learning_in_feedforward_neural_networks} & Analysis of the evolution of the weights of a neural network during training using Mapper & (3) & - \\
\textit{Topological uncertainty: Monitoring trained neural networks through persistence of activation graphs} \citep{topological_uncertainty} & Uncertainty measurement of neural network predictions using the topology of neuron activations & (3) & (3, 5) \\
\textit{Topological measurement of deep neural networks using persistent homology} \citep{topological_measurement_of_deep_networks} & Computation of persistence diagrams using neuron path relevance and connection with network expressivity and problem difficulty & (3) & - \\
\textit{Evaluating the disentanglement of deep generative models with manifold topology} \citep{zhou2021evaluating} & Measurement of disentanglement of generative neural networks & (2) & - \\
\textit{Quantitative performance assessment of CNN units via topological entropy calculation} \citep{zhao2022quantitative} & Measurement of quality of convolutions in a neural network using persistent homology & (3) & - \\
\textit{Overfitting measurement of deep neural networks using no data} \citep{overfitting_measurement_CNNs_using_trained_network_weights} & Study of overfitting by analysing the points near the diagonal of a persistence diagram generated from the weights of a neural network & (3) & - \\
\textit{An adversarial robustness perspective on the topology of neural networks} \citep{goibert2022an} & Analysis of adversarial examples by means of the topology of the subgraph of under-optimized edges of neural networks & (3) & (3) \\
\textit{Representation topology divergence: A method for comparing neural network representations} \citep{representation_topology_divergence} & Definition of similarity between data representations using persistent homology & (3) & (7) \\
\textit{On the topological expressive power of neural networks} \citep{on_the_topological_expressive_power_of_neural_networks} & Measurement of expressivity of network architectures & (2) & - \\
\textit{Generalization bounds using data-dependent fractal dimensions} \citep{generalization_bounds_using_data_dependent} & Data-dependent generalization error bounds using persistent homology dimension on the training weights & (3, 4) & - \\
\textit{TopoBERT: Exploring the topology of fine-tuned word representations} \citep{topobert} & Mapper graph of transformer-based models with applications in finetuning of language models & (3) & - \\
\textit{Experimental observations of the topology of convolutional neural network activations} \citep{experimental_observations_of_the_topology_of_cnns} & Analysis of the Mapper graph of neuron activations and definition of a similarity function between layers using sliced Wasserstein distances between persistence diagrams generated from them & (3) & - \\
\textit{ReLU neural networks, polyhedral decompositions, and persistent homology} \citep{relu_neural_networks_polyhedral_decompositions} & Detection of homological features in manifolds embedded in the input space of a neural network using the polyhedra decomposition induced by a ReLU feedforward neural network & (2) & - \\
\textit{Caveats of neural persistence in deep neural networks} \citep{caveats_of_neural_persistence_in_deep_neural_networks} & Connection between neural persistence and variance measures of neural network weights & (3) & - \\
\textit{On the use of persistent homology to control the generalization capacity of a neural network} \citep{on_the_use_of_ph_to_control_the_generalization_capacity_of_a_nn} & Regression of generalization gap using the average persistence of zero dimensional persistence diagrams & (3) & (6) \\
\textit{Visualizing and analyzing the topology of neuron activations in deep adversarial training} \citep{zhou2023visualizing} & Analysis of the Mapper graph of neuron activations of normally and adversarially trained neural networks & (3) & - \\
\textit{TopP\&R: Robust Support Estimation Approach for Evaluating Fidelity and Diversity in Generative Models} \citep{kim2023toppr} & Approximation of the precision and recall scores for generative models using a persistent homology-based approach.  & (2) & (7)\\
\textit{Topological structure of complex predictions} \citep{gtda_for_decision_regions} & Analysis of the GTDA Reeb network of output spaces of neural networks  & (2) & - \\
\textit{Topological dynamics of functional neural network graphs during reinforcement learning} \citep{topological_dynamics_of_neural_networks_during_rl} & Study of reinforcement learning neural networks during training and inference using homology & (3) & - \\
\end{longtable}